\documentclass[conference]{IEEEtran}
\usepackage{times}
\usepackage{amsmath}
\usepackage{amssymb}
\usepackage{booktabs}
\usepackage{amssymb}
\usepackage{graphicx} 
\usepackage{pifont}    
\usepackage{multirow}  
\usepackage{array}     
\usepackage{cuted}
\usepackage{caption} 
\usepackage{cite}
\usepackage{float}
\usepackage{makecell}
\newcommand{\cmark}{\ding{51}}
\newcommand{\xmark}{\ding{55}}
\usepackage[numbers]{natbib}
\usepackage{multicol}
\usepackage[bookmarks=true]{hyperref}
\usepackage{graphicx}
\graphicspath{{images/}}

\usepackage{booktabs}       
\usepackage{colortbl}       
\usepackage[table]{xcolor}
\usepackage{siunitx}
\usepackage{array}
\usepackage{multirow}       
\definecolor{sectiongray}{gray}{0.90}
\usepackage{amsmath,amssymb}
\usepackage{fontawesome5}

\begin{document}

\title{Beyond Isolation: A Unified Benchmark for General-Purpose Navigation
}

\author{
  Samson Sun$^{1,2,\dagger}$\hspace{1em} Tianyi Yang$^{1,\dagger}$\hspace{1em} Tengyue Wang$^{1,\dagger}$\hspace{1em} Yikai Xue$^{1,\dagger}$\hspace{1em} Zhengjie Xu$^{1,\dagger}$\\
  Lingming Zhang\hspace{1em} Qichen Zhang$^{2}$\hspace{1em} Chao Liang$^{2}$\hspace{1em} Zhipeng Zhang$^{1,2,}$\textsuperscript{\faEnvelope}\\
  $^{1}$AutoLab, SAI, Shanghai Jiao Tong University\hspace{1em} $^{2}$Research Lab, Anyverse Dynamics\hspace{1em}
}



%

\maketitle
\renewcommand{\thefootnote}{}
\footnotetext{$^{\dagger}$Equal contribution. \textsuperscript{\faEnvelope} Corresponding author.}

\begin{strip}
    \vspace{-5em}
    \centering
    \includegraphics[width=1.0\textwidth]{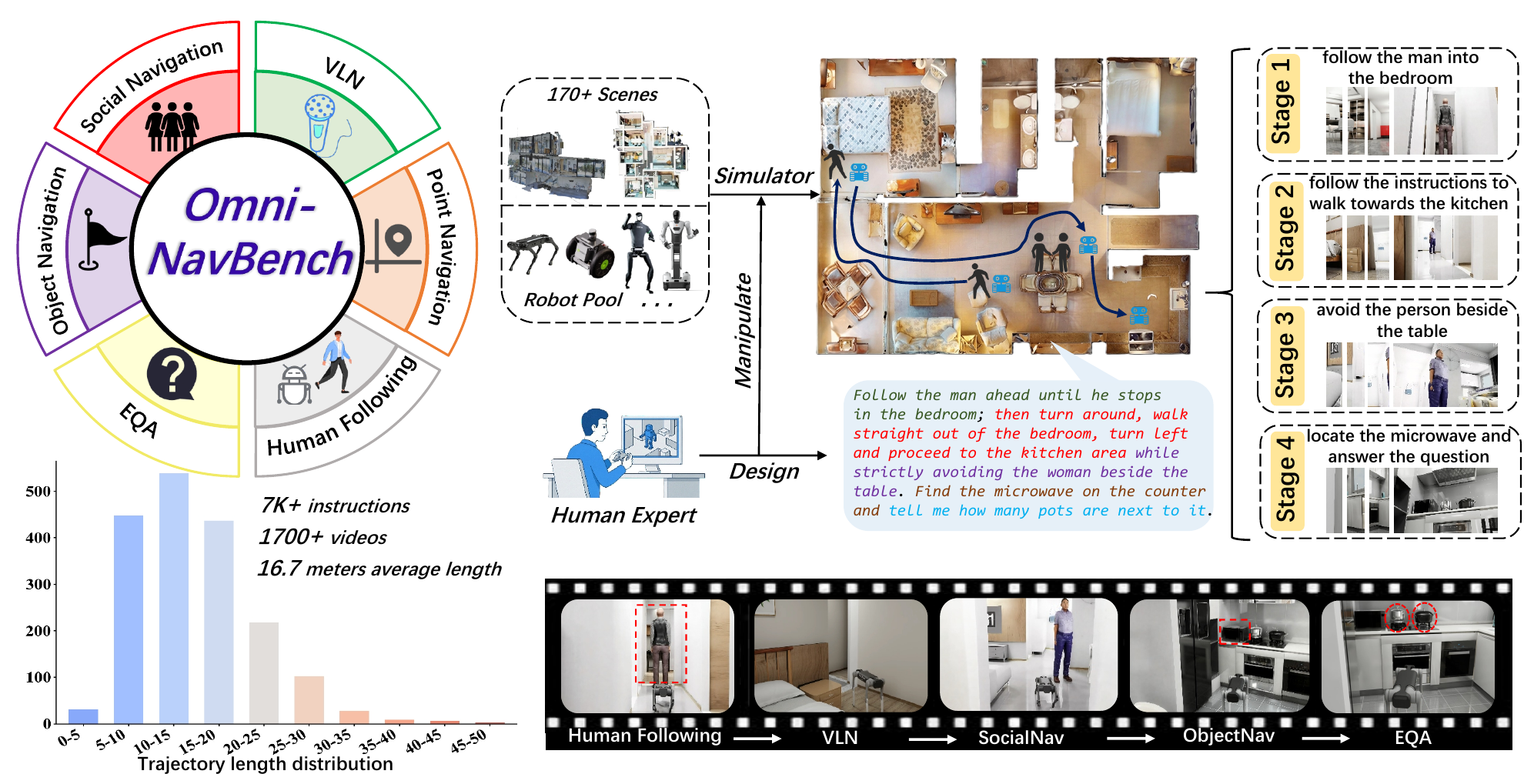}
    \captionof{figure}{\textbf{Overview of the proposed OmniNavBench.} We introduce a unified benchmark designed to evaluate cross-skill coordination and cross-embodiment generalization in embodied navigation. The benchmark constructs composite instructions by dynamically composing a sequence of primary sub-tasks (\textit{i.e.},  VLN, PointNav, ObjectNav, Human Following), which must be executed while concurrently satisfying overarching constraints for SocialNav and EQA.}
    \label{fig:overview}
    \vspace{-0.5em}
\end{strip}

\IEEEpeerreviewmaketitle


\begin{abstract}
The pursuit of general-purpose embodied agents is currently hindered by fragmented evaluation protocols that isolate navigation skills and fixate on specific robot morphologies. This disconnect fails to reflect real-world scenarios where agents must orchestrate diverse behaviors across varying physical embodiments. To bridge this gap, we introduce OmniNavBench, a holistic benchmark designed to rigorously assess cross-skill coordination and cross-embodiment generalization. Distinguished from existing datasets, OmniNavBench introduces three paradigm shifts: \textit{(1) Compositional Complexity.} We propose composite instructions that interleave sub-tasks from 6 distinct categories (\textit{i.e.}, PointNav, VLN, ObjectNav, SocialNav, Human Following and EQA), compelling agents to seamlessly transition between exploration, interaction, and social compliance within a single unified episode. \textit{(2) Morphological Universality and Sensor Flexibility.} We present a simulation platform that breaks the reliance on single-morphology evaluation. This ecosystem empowers researchers to test generalization across different robot types, including humanoid, quadrupedal, and wheeled, while accommodating diverse algorithmic needs through a modular sensor interface and a hybrid suite of \textbf{170} environments blending synthetic assets with real-world scans. \textit{(3) Demonstrations Quality.} Moving beyond mechanical shortest-path algorithms, we curate \textbf{1,779} expert trajectories via human teleoperation, capturing critical behavioral nuances, such as exploratory glance and anticipatory avoidance, essential for natural human-robot coexistence. Extensive evaluations demonstrate that current methods, despite their claimed unified design, struggle to adapt to the complex, interleaved nature of truly general-purpose navigation. This exposes a critical disparity between existing capabilities and the demands of real-world deployment, underscoring OmniNavBench as a crucial testbed for the next generation of generalist navigators. Dataset, code, and leaderboard are available at \url{http://omninavbench.cloud-ip.cc}.
\end{abstract}

\section{Introduction}

Embodied navigation stands as a cornerstone of embodied AI and aims to endow agents with the autonomy to traverse unfamiliar physical environments through perception and reasoning. While the field began with pioneering tasks like Vision-and-Language Navigation (VLN) \cite{anderson2018vision,wei2025streamvln} and
point goal navigation (PointNav) \cite{anderson2018evaluation,zhao2021surprising}, it has subsequently expanded into a diverse array of specialized formulations. These range from PointNav and VLN to more complex behaviors such as object goal navigation (ObjectNav) \cite{chaplot2020object, gao2023room}, social navigation (SocialNav) \cite{biswas2022socnavbench,schreiter2025thor}, human following \cite{huang2017robust, zhang2021efficient,wang2025trackvla}, and embodied question answering (EQA) \cite{majumdar2024openeqa, yu2019multi}. Although recent efforts have sought to develop unified models capable of addressing multiple tasks within a single architecture \cite{zhang2024uni,zhang2025embodied,zheng2024towards,zhou2025same}, the evaluation landscape remains fragmented.

This fragmentation arises because different navigation tasks are typically assessed using isolated benchmarks, where VLN relies on R2R \cite{anderson2018vision} and RxR \cite{ku2020room} while ObjectNav and EQA utilize HM3D-OVON \cite{yokoyama2024hm3d} and OpenEQA \cite{majumdar2024openeqa} respectively. Such a setup creates artificial boundaries between skills by implicitly assuming that agents perform only one type of task at a time. Consequently, models claiming general navigation~\cite{zhang2024uni,zhang2025embodied,zheng2024towards,zhou2025same} capabilities must undergo testing across disparate environments that fail to reflect the reality of deployment where agents must seamlessly switch between diverse behaviors like exploratory search and obstacle avoidance. Furthermore, current evaluation protocols often fail to capture the complexity of physical deployment at both the behavioral and embodiment levels. Most datasets rely on trajectory data generated by shortest-path algorithms \cite{krantz2020beyond, wang2023scaling} which miss the nuances of natural human motion under uncertainty such as exploratory behaviors or anticipatory avoidance \cite{yokoyama2024hm3d, dong2025ha, lee2024citynav, qiu2025egocognav}. This limitation is compounded by evaluations restricted to limited robot morphologies \cite{wang2025rethinking}, meaning that results derived from a single configuration are insufficient to predict performance across the heterogeneous kinematic properties of different embodiments required for real-world applications.

To bridge the gap between fragmented academic tasks and the holistic demands of real-world operation, we introduce \textbf{OmniNavBench}. Particularly, our OmniNavBench makes the following efforts for developing general navigation:

\textbf{\textit{(1) Cross-Skill Coordination Evaluation:}}
We introduce a rigorous protocol to bridge isolated navigation skills, pairing 1,779 composite instructions with expert trajectories to compel seamless transitions between capabilities. This benchmark is augmented to 7,116 instructions with diverse linguistic variations to test robustness against natural language ambiguity. Each instruction weaves together a sequence of at least two primary tasks (\textit{i.e.}, VLN, ObjectNav, PointNav, and HumanFollow), which must be executed while concurrently satisfying overarching constraints for SocialNav and EQA. This complex formulation ensures agents are evaluated on their ability to maintain spatiotemporal continuity while switching strategies, rather than on performance in isolated subtasks.


\textbf{\textit{(2) Flexible and Unified Evaluation Platform:}} We present a high-fidelity simulation platform built upon NVIDIA Isaac Sim \cite{NVIDIA_Isaac_Sim} that breaks the reliance on single-morphology evaluation by enabling generalization tests across humanoid, quadrupedal, and wheeled robots. To accommodate diverse algorithmic needs, the system features a highly modular interface for customizing sensor count, position, and orientation (\textit{e.g.}, RGB-D cameras, LiDAR, panoramic vision systems). This is supported by a hybrid suite of 170 environments, blending 85 high-fidelity synthetic assets from GRScenes \cite{wang2024grutopia} with 85 photorealistic real-world scans from Matterport3D \cite{chang2017matterport3d}.


\textbf{\textit{(3) Human Expert Demonstrations:}} We move beyond the limitations of mechanical shortest-path algorithms by curating a comprehensive dataset of 1,779 expert trajectories collected through human teleoperation. These demonstrations, which span an average length of 16.7 meters and a cumulative distance of 29.5 kilometers, encapsulate the subtle decision-making patterns and hesitation behaviors that characterize natural human navigation. The dataset includes rich egocentric RGB-D video sequences totaling 24 hours and 2.6M frames. By providing these high-quality supervision signals, we offer a resource that enables models to learn human-intuitive strategies for exploration and obstacle avoidance rather than merely optimizing geometric efficiency.

To probe the limits of existing unified navigation models, we conducted a rigorous evaluation on OmniNavBench. Our analysis indicates a substantial performance degradation when models confront complex, multi-stage tasks. Although they can often make progress on individual sub-tasks, they frequently fail to complete the full composite instruction. This suggests the principal challenge lies in sequential task coordination rather than isolated skill deficiency. This weakness extends to dynamic scenarios, where models display inadequate social awareness and struggle to maintain appropriate interaction with moving humans. Furthermore, we observe significant generalization issues, where performance is inconsistent across different robot morphologies, deteriorates in photorealistic scans compared to synthetic environments, and is highly sensitive to variations in instruction style.


In summary, our main contributions are: $\spadesuit$ We introduce OmniNavBench, a unified benchmark for evaluating general-purpose navigation. $\spadesuit$ We build a scalable simulation platform that supports heterogeneous robot embodiments, customizable sensor configurations, and multiple action interfaces, enabling standardized evaluation across embodiment and control settings. $\spadesuit$ We conduct systematic evaluations and in-depth analyses of state-of-the-art models with unified navigation potential under a unified protocol, providing a strong reference for future unified navigation research.

\section{Related Works}

\subsection{Embodied Navigation Benchmarks}
Early embodied navigation research primarily treated tasks as isolated problems, which created a fragmented evaluation landscape. For instance, standard benchmarks like R2R~\cite{anderson2018vision} and RxR~\cite{ku2020room} focus on VLN, while HM3D-OVON \cite{yokoyama2024hm3d} targets ObjectNav. HA-VLN \cite{dong2025ha} introduces dynamic human activities, which is still confined to the VLN domain. This separation prevents us from verifying whether an agent can coordinate multiple skills within a unified task flow. To address these limitations, recent works have introduced more complex settings. LHPR-VLN \cite{song2025towards} extends task horizons with multi-step goals, but these are largely repetitions of homogeneous subtasks that do not require strategy switching. Other benchmarks like GOAT-Bench \cite{khanna2024goat} involve navigating to goals specified by different modalities. Several similar works attempted to unify multiple navigation tasks, including OctoNav~\cite{gao2025octonav} and VLNVerse~\cite{lin2025vlnversebenchmarkvisionlanguagenavigation}. However, their scope is limited as they operate primarily in static settings, failing to incorporate dynamic entities like humans. This prevents them from testing an agent's critical ability to switch strategies in response to real-time events. Furthermore, this issue is compounded by their reliance on trajectory data from shortest-path algorithms. Such simplistic paths do not reflect the complex, nuanced movements of human experts, which is a key feature captured by the manually collected data in our benchmark.

\subsection{Trajectory Generation and Human Demonstrations}

Most existing navigation benchmarks rely heavily on map-based shortest-path algorithms to generate trajectories \cite{dong2025ha, gao2025octonav, song2025towards}. These ``oracle-view" paths exhibit idealized behavior \cite{anderson2018vision, krantz2020beyond} and lack the natural characteristics humans display when facing uncertainty, such as exploration, anticipatory avoidance, and active scanning \cite{yokoyama2024hm3d, dong2025ha, lee2024citynav, qiu2025egocognav}. This distributional gap prevents models from learning human-intuitive strategies. To address this, some works have explored incorporating human demonstration data. For instance, RxR \cite{ku2020room} provides human follower trajectories with temporally aligned language grounding. Similarly, analysis in CityNav \cite{lee2024citynav} reveals that human paths differ significantly from shortest paths, particularly in landmark utilization (36.3\% vs. 24.6\%). The necessity of human data is further supported by research in robot manipulation, such as Robomimic \cite{mandlekar2021matters} and RH20T \cite{fang2023rh20t}. These studies demonstrate that even limited high-quality human demonstrations can capture behavioral diversity that algorithmically generated data cannot replicate. Despite these findings, this paradigm has not yet been adopted in multi-task navigation benchmarks. Current works either persist with shortest-path generation or limit human data collection to single task types \cite{biswas2022socnavbench}, leaving a lack of expert demonstrations that cover complex task flows.


\subsection{Unified Navigation Models}
At the algorithmic level, significant progress has been made toward building unified navigation capabilities. Several works have successfully unified multiple tasks into a single framework. For example, NaviLLM \cite{zheng2024towards} and SAME \cite{zhou2025same} formulate navigation as language generation or modular decision problems. Similarly, UniGoal \cite{yin2025unigoal} and OmniNav \cite{xue2025omninav} manage multi-modal goals and diverse task types within one system, while Uni-NaVid \cite{zhang2024uni} further integrates VLN, ObjectNav, EQA, and HumanFollow into a single architecture. These efforts demonstrate that current model architectures possess strong potential for cross-task generalization. In parallel, there is a growing trend toward developing generalist algorithms that adapt to different robot morphologies. NavFoM \cite{zhang2025embodied} has been deployed across diverse platforms, including quadrupeds, drones, and wheeled robots. InternVLA-N1 \cite{wei2025ground} showcases zero-shot cross-embodiment generalization, spanning wheeled to humanoid robots, and NaVILA \cite{cheng2024navila} achieves transfer for legged robots by decoupling high-level planning from low-level control. Furthermore, VLN-PE \cite{wang2025rethinking} highlights the importance of physical embodiment, revealing how sensitive existing algorithms are to robot dynamics and camera configurations. Collectively, these advances indicate that unified navigation models have demonstrated promising capabilities in task generalization and continuous control. However, a critical gap remains. These sophisticated models are still evaluated on fragmented, single-task benchmarks. There is currently no unified platform capable of systematically verifying their ability to perform dynamic strategy switching across skills and adapt to different embodiments within a cohesive task flow.

\section{OmniNavBench}
The goal of OmniNavBench is to offer a unified platform for evaluating cross-skill navigation capabilities across heterogeneous robot embodiments, utilizing high-quality human demonstrations as behavioral references. To achieve this, we adhere to the following four core design principles.

\noindent $\spadesuit$ \textbf{Compositional Task Structure.} To evaluate the coordination of multiple navigation skills, each instruction integrates at least two heterogeneous sub-tasks, which combine static goal reaching, dynamic target tracking, and semantic reasoning within a single episode. This design imposes spatio-temporal continuity constraints, necessitating adaptive strategy switching as task demands evolve.

\noindent $\spadesuit$ \textbf{Cross-Embodiment Coverage.} Addressing the diversity of real-world platforms, we ensure the same instruction set is executable across three morphologically distinct agents, including wheeled, quadrupedal, and humanoid robots. This design enables a direct assessment of algorithmic robustness against variations in kinematics, viewpoints, and dynamics.

\noindent $\spadesuit$ \textbf{Naturalistic Behavior Supervision.} Prioritizing behavioral realism over geometric optimality, all reference trajectories are collected via human teleoperation. Unlike shortest-path algorithms (\textit{e.g.}, A*), this approach preserves human-specific navigation patterns, such as exploratory behaviors and anticipatory adjustments \cite{yokoyama2024hm3d, dong2025ha, lee2024citynav, qiu2025egocognav}.

\noindent $\spadesuit$ \textbf{Controlled Difficulty Levels.} To facilitate systematic evaluation, we establish a complexity curriculum by modulating the sequence length and task heterogeneity. The benchmark offers a difficulty gradient, ranging from two-task compositions to long-horizon four-or-more-task sequences, which allows for fine-grained analysis of agent capabilities.

\subsection{Task Specification}

We formalize the navigation mission as a composite instruction comprising an ordered sequence of sub-tasks defined by

\begin{equation}
    \mathcal{M} = \{(l_1, \tau_1, g_1), (l_2, \tau_2, g_2), \ldots, (l_n, \tau_n, g_n)\}
    \label{eq:energy}
\end{equation}
where $l_i$ represents the natural language instruction segment, $\tau_i$ specifies the sub-task type, and $g_i$ defines the target goal. Our benchmark incorporates the following primitives to construct these sequences.

\noindent $\spadesuit$ \textbf{Sequential Sub-tasks.} We integrate four distinct navigation primitives to evaluate a full spectrum of embodied capabilities.

\begin{itemize}

\item \textbf{VLN} requires the agent to interpret natural language instructions grounded in visual landmarks, such as \textit{``turning right after passing a sofa''.} 

\item \textbf{ObjectNav} compels the agent to conduct semantic exploration in unknown environments to locate instances of specific categories like \textit{``find a trash can''.} 

\item \textbf{PointNav} tests precise geometric understanding by requiring movement to exact metric coordinates $(x, y)$ relative to the starting pose. 

\item \textbf{HumanFollow} necessitates active visual tracking where the agent must maintain continuous line-of-sight with a moving human subject.

\end{itemize}

Notably, \textbf{Return} is a VLN sub-task that mandates navigating back to the starting position upon task completion, requiring the agent to leverage long-term spatial memory.

\noindent $\spadesuit$ \textbf{Parallel Sub-task.} We designate \textbf{SocialNav} and \textbf{EQA} as a concurrent operational constraint capable of being superimposed onto any of the aforementioned sequential sub-tasks. 


\begin{itemize}
    \item \textbf{SocialNav} requires the agent to minimize intrusion into human activity spaces during navigation.

    \item \textbf{EQA} fuses navigation with cognitive reasoning where the agent must gather visual evidence to answer questions regarding environmental attributes.
\end{itemize}


\begin{figure}[!t] 
    \centering
    \includegraphics[width=\linewidth]{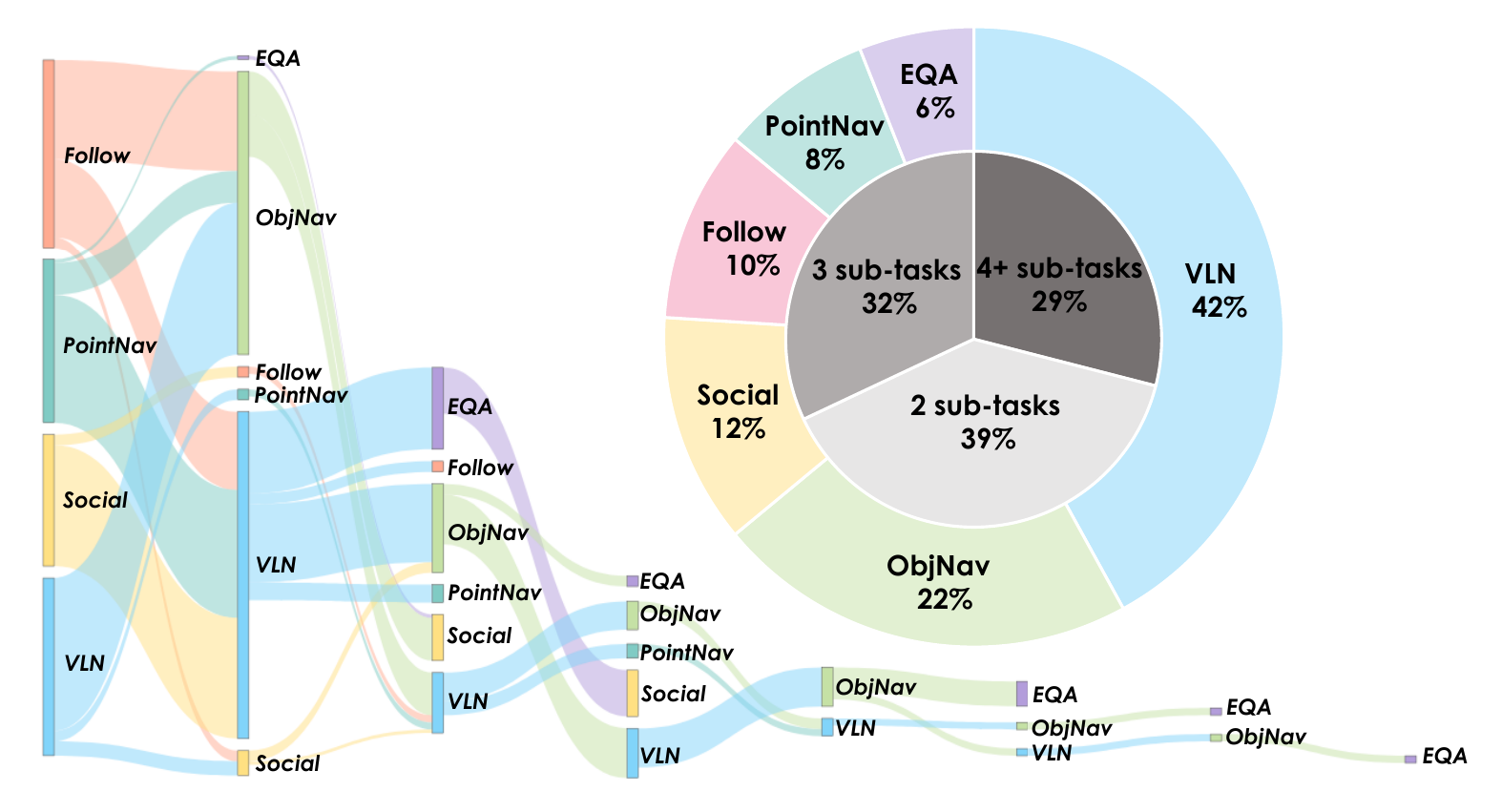} 
    \caption{\textbf{Statistics of Sub-instructions.} 
    Sankey diagram depicting cross-skill compositional patterns (left), and distribution of sub-instruction types with task complexity breakdown (right). }
    \label{fig:subInstructions}
    \vspace{-1.5em}
\end{figure}

\subsection{Data Acquisition}

We adhere to a strict human teleoperation protocol to capture the nuances of biological navigation rather than relying on algorithmic planners. While shortest-path algorithms prioritize geometric optimality, human controllers exhibit organic behaviors like hesitation and exploratory glances \cite{yokoyama2024hm3d, dong2025ha, lee2024citynav, qiu2025egocognav}. These subtle cues are essential for learning robust policies, particularly for dynamic tasks like HumanFollow where rule-based generators fail to produce convincing social compliance.

\noindent $\spadesuit$ \textbf{Rigorous Collection Pipeline.} We implemented a multi-stage workflow involving trained annotators to ensure high-fidelity data. Operators manually control the robot to execute composite instructions within the simulator while the system records pose and sensor data as waypoints. To guarantee semantic alignment, experts subsequently verify the trajectories by reviewing the egocentric video feeds. We mandate that any trajectory failing to capture specified landmarks or objects clearly must be revised by the original operator to maintain behavioral consistency throughout the iteration loop.

\noindent $\spadesuit$ \textbf{Morphology-Agnostic Representation.} The final dataset comprises waypoint sequences $(x, y, \theta, t)$ paired with synchronized RGB-D streams. We deliberately exclude low-level action primitives because the high-fidelity physics in Isaac Sim introduces inertia and friction that necessitate continuous micro-corrections. These task-irrelevant adjustments constitute noise for high-level policy learning whereas our waypoint representation abstracts away these dynamics to facilitate transfer across different robot controllers.

\begin{table*}[t]
\centering
\caption{\textbf{Comparisons between OmniNavBench and existing benchmarks.} OmniNavBench supports the most comprehensive set of tasks and utilize human-expert trajectories.}
\label{tab:comparison}
\resizebox{\textwidth}{!}{%
\begin{tabular}{l c c c c c c c c c c}
\toprule
\multirow{2}{*}{\textbf{Benchmark}} & \multicolumn{7}{c}{\textbf{Task Capability}} & \textbf{Cross} & \textbf{Trajectory} & \textbf{Avg. Instruction} \\
\cmidrule(lr){2-8} 
 & \textbf{ObjNav} & \textbf{SocialNav} & \textbf{VLN} & \textbf{Follow} & \textbf{PointNav} & \textbf{EQA} & \textbf{Unified} & \textbf{Embodiment} & \textbf{Source} & \textbf{Length} \\
\midrule
R2R \cite{anderson2018vision} & \xmark & \xmark & \cmark & \xmark & \xmark & \xmark & \xmark & \xmark & Algorithm & 29 \\
RxR \cite{ku2020room} & \xmark & \xmark & \cmark & \xmark & \xmark & \xmark & \xmark & \xmark & Human Expert & - \\
VLN-CE \cite{krantz2020beyond} & \xmark & \xmark & \cmark & \xmark & \xmark & \xmark & \xmark & \xmark & Algorithm & 30 \\
OpenEQA \cite{majumdar2024openeqa} & \xmark & \xmark & \xmark & \xmark & \xmark & \cmark & \xmark & \xmark & - & - \\
HA-VLN \cite{dong2025ha} & \xmark & \cmark & \cmark & \xmark & \xmark & \xmark & \xmark & \xmark & Algorithm &  - \\
GOAT-Bench \cite{khanna2024goat}& \cmark & \xmark & \xmark & \xmark & \xmark & \xmark & \xmark & \xmark & Algorithm & - \\
REVERIE \cite{qi2020reverie} & \xmark & \xmark & \cmark & \xmark & \xmark & \xmark & \xmark & \xmark & Algorithm & 18 \\
OVON \cite{yokoyama2024hm3d} & \cmark & \xmark & \xmark & \xmark & \xmark & \xmark & \xmark & \xmark & Algorithm & - \\
LHPR-VLN \cite{song2025towards}& \cmark & \xmark & \cmark & \xmark & \xmark & \xmark & \cmark & \cmark & Algorithm & 18.17 \\
VLN-PE \cite{wang2025rethinking} & \xmark & \xmark & \cmark & \xmark & \xmark & \xmark & \xmark & \cmark & Algorithm & - \\
VLNVerse \cite{lin2025vlnversebenchmarkvisionlanguagenavigation} & \cmark & \xmark & \cmark & \xmark & \xmark & \xmark & \cmark & \cmark & Algorithm & - \\
OctoNav-Bench \cite{gao2025octonav}& \cmark & \xmark & \cmark & \xmark & \cmark & \xmark & \cmark & \xmark & Algorithm & - \\
\midrule
\textbf{OmniNavBench (Ours)} & \textbf{\cmark} & \textbf{\cmark} & \textbf{\cmark} & \textbf{\cmark} & \textbf{\cmark} & \textbf{\cmark} & \textbf{\cmark} & \textbf{\cmark} & \textbf{Human Expert} & 42 \\
\bottomrule
\end{tabular}%
}
\vspace{-1em}
\end{table*}
\begin{figure}[!t] 
    \centering
    \includegraphics[width=\linewidth]{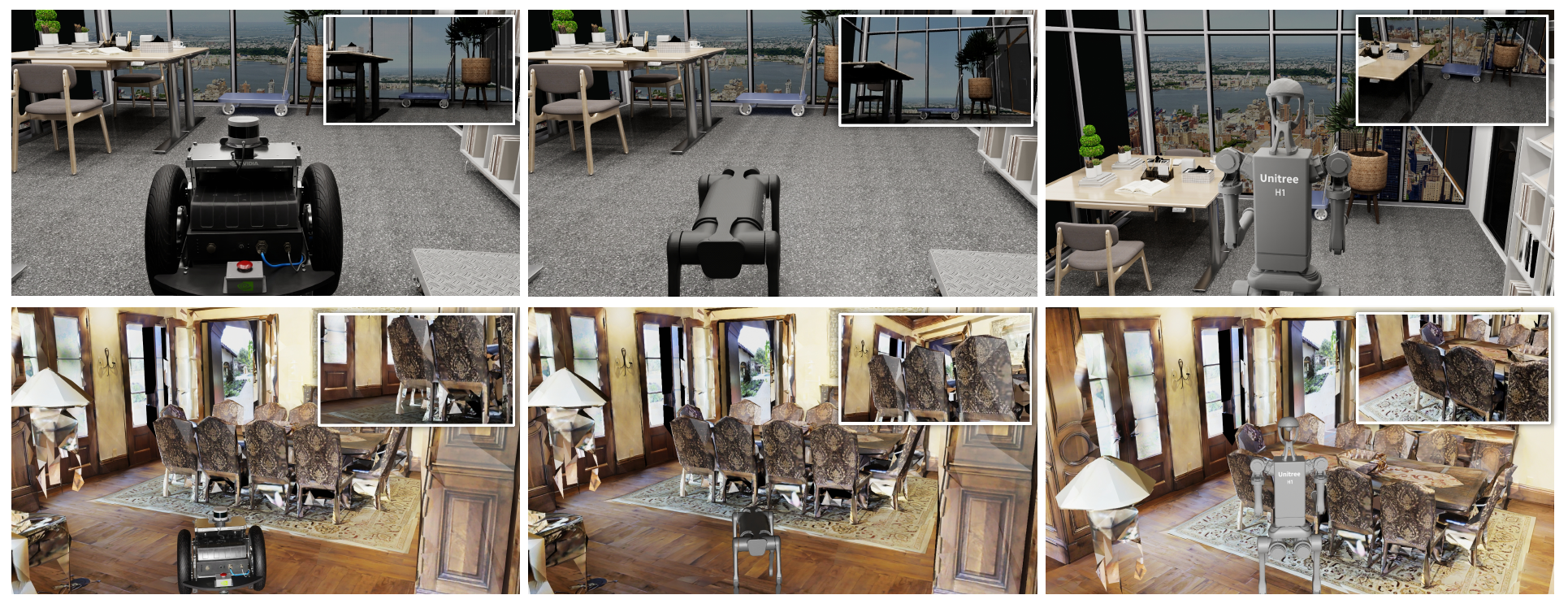} 
    \caption{\textbf{Cross-embodiment navigation across diverse environments.} Each row shows a different robot morphology (Carter, Aliengo, H1); each column shows a different environment source (synthetic, Matterport3D). Insets display egocentric RGB observations, highlighting viewpoint variations across embodiments.}
    \label{fig:cross_embodiment}
    \vspace{-2em}
\end{figure}

\subsection{Dataset Statistics}

OmniNavBench contains 1,779 expert demonstration trajectories spanning 170 distinct scenes accompanied by 7,116 corresponding instructions, with an average trajectory length of 16.7 meters and duration of 48.9 seconds. The scenes are drawn from three sources with different visual characteristics: home environments account for 828 trajectories (46.5\%), covering diverse residential layouts and furniture arrangements; commercial spaces contribute 213 trajectories (12.0\%), including offices and retail environments; and Matterport3D scans provide 738 trajectories (41.4\%) with photorealistic reconstructions of real-world buildings. This diversity in visual appearance and spatial structure enables evaluation of cross-domain generalization.

\noindent $\spadesuit$ \textbf{Task and Linguistic Complexity.} Fig. \ref{fig:subInstructions} illustrates the comprehensive statistics of our dataset. The task composition (outer ring) demonstrates a balanced distribution covering both static and dynamic objectives. Specifically, VLN accounts for the largest share at 42\%, followed by ObjectNav at 22\%. To ensure diversity in interaction, we incorporate significant proportions of SocialNav (12\%), Follow (10\%), PointNav (8\%), and EQA (6\%). Notably, we allocated 28.6\% of the VLN category to Return episodes, a deliberate design choice to rigorously evaluate the agent's long-term spatial memory and path integration capabilities as previously discussed. The complexity distribution (inner ring) reveals a deliberate gradient of difficulty designed for systematic evaluation. Specifically, 39\% of trajectories contain 2 sub-tasks, 32\% extend to 3 sub-tasks, and a substantial 29\% involve sequences of 4 or more sub-tasks.The instructions exhibit rich linguistic diversity, encompassing spatial references (\textit{e.g.}, bedroom, kitchen), target objects (\textit{e.g.}, chair, sofa), and action verbs (\textit{e.g.}, forward, follow), with lengths spanning from 9 to over 130 words (see Appendix Fig.~\ref{fig:length_appendix}).

\noindent $\spadesuit$ \textbf{Instruction and Linguistic Diversity} Our benchmark further provides three supplementary instruction variants which are \textit{Concise}, \textit{First Person}, and \textit{Verbose} to enhance linguistic diversity, yielding a total of 7,116 stylized instructions. These variants were generated using Qwen3-Max~\cite{qwen3} to produce semantically consistent paraphrases of the original instructions. \textit{Concise} instructions retain only essential navigation intents by removing redundant descriptions. \textit{First Person} instructions reframe the original into a first-person voice. \textit{Verbose} instructions incorporate additional contextual elaborations while preserving the target and action sequence. Details about stylized instructions are shown in the appendix due to space limitations.

\noindent $\spadesuit$ \textbf{Morphology and Evaluation Protocols.} The benchmark encompasses three robot morphologies including wheeled (NVIDIA Carter V1 and Kitt15), quadrupedal, and bipedal models where each category accounts for approximately one-third of the trajectories. Notably, the quadrupedal subset specifically includes stair-climbing sequences that are largely absent from the other two groups. To facilitate rigorous assessment, the trajectory split partitions the dataset by scene and contains 1,187 training trajectories across 112 scenes along with 592 test trajectories from 58 scenes to evaluate generalization to novel environments. 


\begin{table*}[!t]
\centering
\caption{\textbf{Sub-task Definitions and Completion Rules in OmniNavBench.}
A composite instruction consists of sub-tasks $\mathcal{M}=\{(l_i, \tau_i, g_i)\}$.
For each $\tau_i$, we define its input $l_i$, evaluator goal $g_i$, and the sub-task completion rule.}
\label{tab:task_definitions}
\resizebox{\textwidth}{!}{%
\begin{tabular}{l|l|l|l}
\toprule
\textbf{Task Type ($\tau_i$)} &
\textbf{Instruction Input ($l_i$)} &
\textbf{Goal ($g_i$)} &
\textbf{Completion Rule} \\
\midrule

\multicolumn{4}{l}{\color{gray}\textit{\textbf{State-Goal Navigation} (Agent must reach a specific terminal state)}} \\
\textbf{VLN} &
Language with Landmarks \& Rooms &
Landmarks set $\mathcal{L}_{goal}$ and Region $\mathcal{R}_{goal}$ &
$\exists t:\; \big(\min_{\ell\in \mathcal{L}_{goal}} d_{geo}(x_t, \ell)\le 3.0\,\text{m}\big)\;\land\;\big(x_t\in \mathcal{R}_{goal}\big)$ \\
\textbf{PointNav} &
Relative Displacement $\Delta p$ &
Evaluator goal coordinate $p_{goal}$ &
$\exists t:\; d(x_t, p_{goal}) \le 0.36\,\text{m}$ \\
\textbf{ObjectNav} &
Open-vocabulary Query (text) &
Target object set $\mathcal{O}(\text{query})$ &
$\exists t:\; \min_{o\in \mathcal{O}(\text{query})} d_{geo}(x_t, o)\le 1.0\,\text{m}$ \\

\midrule

\multicolumn{4}{l}{\color{gray}\textit{\textbf{Process-Maintenance} (Agent must maintain a state over time)}} \\
\textbf{HumanFollow} &
``Follow [Description]'' &
Moving person $H_{id}$ &
$\text{Cond}_{t_e} = \text{True}$, where $\text{Cond}_t := (1.0 < d_t \le 3.0)\land(\angle_{err,t}\le 60^\circ)$ \\

\midrule

\multicolumn{4}{l}{\color{gray}\textit{\textbf{Cognitive Reasoning} (Agent must generate an answer)}} \\
\textbf{EQA} &
Visual Question $Q$ &
Ground Truth Answer $A$ &
$\text{SubstringMatch}(A_{\text{pred}}, A) = \text{True}$ \\

\midrule\midrule

\multicolumn{4}{l}{\color{gray}\textit{\textbf{Global Constraint} (Applies to all tasks)}} \\
\textbf{SocialNav} &
(Implicit) &
All avoided humans $\mathcal{H}$ &
Constraint: $d(x_t, \mathcal{H}) \ge 1.2\,\text{m}$ throughout the episode \\
\bottomrule
\end{tabular}%
}
\vspace{2pt}
\begin{minipage}{\textwidth}
\footnotesize
\textbf{Notation.}
$x_t$: robot center position;
$d_{geo}(\cdot,\cdot)$: geodesic distance computed via NavMesh;
$d(\cdot,\cdot)$: 2D Euclidean distance;
$\angle_{err,t}$: angle between robot heading and robot-to-human vector;
$H_{id}$: identity of the person to follow;
$t_e$: timestep when the person stops moving;
For State-Goal tasks, completion is determined at the \textbf{first} timestep satisfying the predicate.
Episode-level metrics (SR, SPL, \textit{etc.}) are defined in Tab.~\ref{tab:metrics_corrected}.
\end{minipage}
\end{table*}

\begin{table*}[!t]
\centering
\caption{\textbf{Evaluation Metrics in OmniNavBench.} All reported metrics represent the average over test episodes (using the relevant subset $N_{\text{task}}$ for task-specific metrics).
\textbf{SR}, \textbf{SPL}, and \textbf{OSR} are standard navigation metrics.
\textbf{SGC} measures continuous progress.
\textbf{HFS}, \textbf{HFR}, \textbf{ACC}, and \textbf{SII} evaluate human-robot interaction and safety.}
\label{tab:metrics_corrected}

\renewcommand{\arraystretch}{1.2}

\resizebox{\textwidth}{!}{%
\begin{tabular}{l|l|c|l|l}
\toprule
\textbf{Dimension} & \textbf{Metric} & \textbf{Sym.} & \textbf{Definition / Formula (Averaged over relevant episodes)} & \textbf{Significance} \\
\midrule

\multirow{4}{*}{\textbf{Completion}}
& Success Rate & SR & $\frac{1}{N} \sum_{i=1}^{N} S_i(\delta_g)$ & Fraction of fully successful episodes. \\

& Completion Success Rate & CSR & $\frac{1}{N} \sum_{i=1}^{N} \frac{K_{\text{completed},i}}{K_{\text{total},i}}$ & Avg. ratio of strictly completed sub-instructions. \\

& Sub-Goal Completion & SGC & $\frac{1}{N} \sum_{i=1}^{N} \frac{1}{K_i} \sum_{j=1}^{K_i} \mathcal{P}(\tau_{i,j})$ & Average continuous progress across sub-tasks. \\

& Oracle Success Rate & OSR & $\frac{1}{N} \sum_{i=1}^{N} OS_i(\delta_g)$ & Success if goal was \textit{ever} reached. \\
\midrule

\multirow{1}{*}{\textbf{Efficiency}}
& Success weighted by Path Length & SPL & $\frac{1}{N} \sum_{i=1}^{N} S_i \frac{\ell_{shortest,i}}{\max(\ell_{actual,i}, \ell_{shortest,i})}$ & Efficiency weighted by success. \\
\midrule

\multirow{3}{*}{\textbf{Interaction}}
& Human Follow Success & HFS & $\frac{1}{N_{\text{follow}}} \sum_{i=1}^{N_{\text{follow}}} \mathbb{I}(\text{pos}_{\text{end},i} \in \text{Valid}_i)$ & Binary success at the final step. \\

& Human Follow Ratio & HFR & $\frac{1}{N_{\text{follow}}} \sum_{i=1}^{N_{\text{follow}}} \frac{T_{\text{valid},i}}{T_{\text{follow},i}}, \enskip T_{\text{valid},i} = \sum_{t=1}^{T_{\text{follow},i}} \mathbb{I}(\text{Cond}_{i,t})$ & Coverage ratio during following phase. \\

& EQA Accuracy & EQA-ACC & $\frac{1}{N_{\text{eqa}}} \sum_{i=1}^{N_{\text{eqa}}} \mathbb{I}(a_{\text{pred},i} \cong a_{\text{gt},i})$ & Accuracy of numeric-normalized substring match. \\
\midrule

\multirow{1}{*}{\textbf{Safety}}
& Social Intrusion Index & SII & $\frac{1}{N_{\text{social}}} \sum_{i=1}^{N_{\text{social}}} \big( \frac{1}{T_i} \sum_{t=1}^{T_i} \mathbb{I}(d_{t} < 1.2\,\text{m}) \big)$ & Ratio of steps violating personal space. \\

\bottomrule
\end{tabular}%
}
\vspace{2pt}
\begin{minipage}{\textwidth}
\footnotesize
\textbf{Notation.}
$N$: total test episodes ($N_{\text{follow}}, N_{\text{social}}, N_{\text{eqa}}$: task-specific subsets);
$\mathbb{I}(\cdot)$: indicator function;
$K_{\text{total},i}$: total number of sub-instructions in episode $i$;
$K_i$: number of sub-tasks for progress calculation;
$\tau_{i,j}$: the $j$-th sub-task of episode $i$;
$\mathcal{P}(\cdot)$: continuous progress function (see Appendix);
$S_i$: binary success ($S_i=1$ if agent issues \texttt{Stop} within $\delta_g$ of the final goal);
$OS_i$: oracle success ($OS_i=1$ if agent ever reaches within $\delta_g$ of the final goal);
$\delta_g$: goal distance threshold (2.0\,m);
$\ell$: path length;
$T_{\text{follow},i}$: duration of the following phase;
$a_{\text{gt}}, a_{\text{pred}}$: ground truth and predicted answers ($\cong$ denotes a success condition via numeric-normalized substring matching);
$d_{t}$: distance between robot and human, SII threshold of 1.2\,m corresponds to Hall's personal distance zone \cite{hall1966hidden}.
\end{minipage}
\vspace{-1em}
\end{table*}

\subsection{Evaluation Platform}
We established an integrated simulation platform based on NVIDIA Isaac Sim to facilitate unified navigation evaluation and cross-embodiment algorithm development. Distinct from some modern simulators (\textit{e.g.}, Matterport3D~\cite{Matterport3D}, Habitat~\cite{szot2021habitat}) that employ discrete state transitions or teleportation-based navigation, our platform performs continuous physical stepping with full dynamics simulation. We adopt the RL-based locomotion controllers from VLN-PE \cite{wang2025rethinking} to ensure physically grounded robot dynamics. This design provides high-fidelity physics and photorealistic rendering which ensures that evaluated agents must handle realistic motion dynamics, sensor noise, and visual complexity that closely approximate real-world deployment conditions.

Our platform integrates three morphologically distinct robots including the wheeled NVIDIA Carter V1, the quadrupedal Unitree Aliengo, and the humanoid Unitree H1 within a unified framework. Fig.~\ref{fig:cross_embodiment} visualizes the three embodiments navigating in both synthetic and photorealistic environments, highlighting the viewpoint variations arising from different morphologies. Despite fundamental differences in kinematic constraints and viewpoint heights, the platform achieves unified orchestration through abstracted control interfaces. Algorithms can output waypoint sequences for low-level path tracking or directly issue velocity commands $(v, \omega)$ for end-to-end control. Furthermore, the system supports discrete actions such as step forward or rotate left and right that are mapped to physically-simulated movements. These actions reach the same target poses as teleportation-based simulators while preserving realistic dynamics, which enables the same algorithm to be evaluated across different morphologies without modification. Our platform also facilitates modular sensor configuration where users can specify sensor types including RGB, depth, panoramic camera, LiDAR, and IMU along with their quantities and mounting poses via configuration files.

To enable dynamic scene interaction, the platform leverages the built-in human character system of Isaac Sim which offers diverse pedestrian models with various appearances. Pedestrians follow predefined trajectories while maintaining the capability for autonomous obstacle avoidance to create realistic social navigation scenarios for SocialNav and HumanFollow tasks. The platform supports both synthetic environments with controllable lighting and real-world Matterport scans for cross-domain generalization evaluation. To ensure scientific rigor, scene initial states, pedestrian trajectories, and random seeds can be precisely reproduced via configuration files which guarantees verifiable and reproducible experimental results.

\subsection{Comparison to Other Benchmarks} As detailed in Tab. \ref{tab:comparison}, OmniNavBench significantly advances upon previous benchmarks. While recent works like OctoNav-Bench~\cite{gao2025octonav} and VLNVerse~\cite{lin2025vlnversebenchmarkvisionlanguagenavigation} move towards task unification, they are largely confined to static environments and neglect dynamic interaction. In contrast, OmniNavBench uniquely integrates six tasks spanning the full spectrum from static navigation (\textit{e.g.}, PointNav) to dynamic human-agent interaction (\textit{e.g.}, SocialNav). Most critically, OmniNavBench distinguishes itself through its behavioral realism. Unlike virtually competing unified benchmarks (\textit{e.g.}, LHPR-VLN \cite{song2025towards}, VLNVerse \cite{lin2025vlnversebenchmarkvisionlanguagenavigation}, OctoNav-Bench \cite{gao2025octonav}) that rely on geometrically optimal paths from algorithmic planners, it is built entirely on human-expert teleoperated data. This foundation allows it to capture naturalistic nuances like exploratory behaviors and social compliance, which are absent in algorithmic trajectories \cite{yokoyama2024hm3d, dong2025ha, lee2024citynav, qiu2025egocognav}. Furthermore, by incorporating diverse robot morphologies and the highest linguistic complexity (avg. 42 words), OmniNavBench provides a more faithful approximation of real-world challenges.

\section{Experiments}


\begin{table*}[t]
    \centering
    \caption{\textbf{Quantitative comparison with state-of-the-art baselines across different robot embodiments.} All metrics are reported in \%. \textbf{CSR} (Completion Success Rate) and \textbf{SGC} (Sub-Goal Completion) indicate partial progress. \textbf{SII} (Social Intrusion Index) is a safety metric where \textbf{lower is better} ($\downarrow$). \textbf{EQA-ACC} denotes the Embodied Question Answering accuracy. Best results are bolded.}
    \label{tab:main_results}
    
    \footnotesize
    \renewcommand{\arraystretch}{0.92}
    \setlength{\tabcolsep}{4pt}
    
    \sisetup{
        detect-weight=true,
        detect-inline-weight=math,
        table-number-alignment=center
    }

    \begin{tabular}{ll *{9}{c}}
    \toprule
    \textbf{Method} & \textbf{Embodiment} & 
    {SR $\uparrow$} & {CSR $\uparrow$} & {SGC $\uparrow$} & {OSR $\uparrow$} & {SPL $\uparrow$} & {HFS $\uparrow$} & {HFR $\uparrow$} & {SII $\downarrow$} & {EQA-ACC $\uparrow$} \\
    \midrule
    
    \multirow{3}{*}{\textbf{MTU3D}} 
      & NVIDIA Carter   & 0.00 & 1.08 & 29.65 & 5.38 & 0.00 & 20.00 & 24.66 & 4.46 & \multicolumn{1}{c}{--} \\
      & Unitree Aliengo & 0.00 & 0.98 & 30.95 & 11.76 & 0.00 & 11.11 & 16.91 & 5.65 & \multicolumn{1}{c}{--} \\
      & Unitree H1      & 0.00 & 2.08 & 27.42 & 8.33 & 0.00 & 22.22 & 27.83 & 6.74 & \multicolumn{1}{c}{--} \\
    \midrule
    
    \multirow{3}{*}{\textbf{Poliformer}} 
      & NVIDIA Carter   & 0.98 & 0.98 & 29.12 & \textbf{10.78} & 0.73 & 14.81 & 14.70 & \textbf{2.72} & \multicolumn{1}{c}{--} \\
      & Unitree Aliengo & 1.00 & 0.00 & 29.41 & 16.00 & 0.29 & 3.70 & 9.68 & \textbf{2.94} & \multicolumn{1}{c}{--} \\
      & Unitree H1      & \textbf{1.08} & 0.00 & 25.27 & 3.23 & 0.31 & 8.00 & 9.32 & 4.14 & \multicolumn{1}{c}{--} \\
    \midrule
    
    \multirow{3}{*}{\textbf{Uni-NaVid}} 
      & NVIDIA Carter   & 1.94 & \textbf{2.91} & \textbf{38.93} & 9.71 & 1.45 & \textbf{45.16} & \textbf{43.67} & 9.39 & 0.00 \\
      & Unitree Aliengo & 1.96 & \textbf{2.94} & \textbf{44.54} & \textbf{26.47} & 1.56 & \textbf{23.33} & \textbf{32.75} & 18.46 & 6.25 \\
      & Unitree H1      & 1.06 & \textbf{2.13} & \textbf{29.02} & \textbf{8.51} & \textbf{1.06} & \textbf{24.14} & \textbf{25.99} & 10.09 & 7.69 \\
    \midrule
    
    \multirow{3}{*}{\textbf{OmniNav}} 
      & NVIDIA Carter   & \textbf{1.96} & 0.00 & 32.34 & 8.82 & \textbf{1.71} & 33.33 & 31.04 & 5.23 & \multicolumn{1}{c}{--} \\
      & Unitree Aliengo & \textbf{8.74} & 1.94 & 35.30 & 15.53 & \textbf{7.01} & 14.29 & 17.49 & 3.40 & \multicolumn{1}{c}{--} \\
      & Unitree H1      & 0.00 & 0.00 & 24.82 & 2.13 & 0.00 & 14.81 & 13.93 & \textbf{3.31} & \multicolumn{1}{c}{--} \\
    \bottomrule
    \end{tabular}
    \vspace{-1em}
\end{table*}



Successful completion of a composite instruction requires valid execution of all sequential sub-tasks and fulfillment of end-of-task criteria. Task definitions and completion rules are detailed in Tab.~\ref{tab:task_definitions}, with episode-level metrics in Tab.~\ref{tab:metrics_corrected}. We provide more details about the metrics definition in the appendix due to space limitation.


%

\subsection{Evaluating Unified Navigation Models}

We evaluate four recent popular unified navigation models representing different unification mechanisms. PoliFormer \cite{zeng2024poliformer} relies on scaled transformer-based reinforcement learning to induce general navigation behaviors across tasks while Uni-NaVid \cite{zhang2024uni} formulates diverse navigation tasks as a single vision–language–action generation problem from egocentric video. MTU3D \cite{zhu2025move} unifies navigation and exploration by explicitly optimizing movement for spatial understanding and OmniNav \cite{xue2025omninav} adopts a shared continuous waypoint abstraction to support multiple navigation task formulations. All models are evaluated zero-shot with publicly released weights, adapting only camera count, resolution, and FOV to each algorithm's specifications.

\noindent \textbf{\textit{Findings1: Universal Performance Degradation.}} Tab.~\ref{tab:main_results} presents the quantitative evaluation of four unified navigation models across three robot embodiments. Despite their claimed multi-task capabilities demonstrated on isolated benchmarks, all evaluated models exhibit substantial performance degradation on our OmniNavBench. The highest \texttt{Success Rate} (\texttt{SR}), which measures whether the agent issues a stop action at goal area, achieved is merely 8.74\% by OmniNav on the Unitree Aliengo platform while most configurations fall below 2\%. Crucially, this underperformance is consistent across all three morphologies which indicates a fundamental limitation rather than platform-specific issues. Consequently, the finding underscores the necessity of OmniNavBench as a rigorous testbed to bridge the gap between specialized success and true cross-task cross-embodiment generalization.

\noindent \textbf{\textit{Findings2: Disparity Between Isolated Skills and Sequential Execution.}} \texttt{Sub-Goal Completion} (\texttt{SGC}) is defined as the mean progress across sub-tasks, where each sub-task progress score ranges from 0 to 1. The contrast between \texttt{SGC} and the \texttt{Success Rate} (\texttt{SR}) reveals a critical insight where models can achieve moderate progress on individual sub-tasks with \texttt{SGC} exceeding 44\% for Uni-NaVid on Aliengo yet fail to complete full episodes. This disparity indicates that the primary bottleneck lies in the inability to execute multiple diverse navigation tasks sequentially within a single continuous episode. This limitation largely stems from current training paradigms which rely on datasets composed of isolated single task and fail to model the realistic transition between diverse navigation behaviors, which validates the necessity of our proposed dataset for fostering true multi-task sequential capability.




\noindent \textbf{\textit{Findings3: The Last Centimeter Gap in End-to-End Navigation.}} \texttt{Oracle Success Rate} (\texttt{OSR}) captures instances where the agent physically reaches goal once without issuing a stop action. Interestingly, Uni-NaVid achieves 26.47\% \texttt{OSR} on the Unitree Aliengo yet secures only a 1.96\% \texttt{SR}. This significant disparity between higher \texttt{OSR} values and the lower \texttt{SR} suggests that agents frequently navigate correctly to the vicinity of target goals but fail to issue correct termination signals. These results reflect that current end-to-end navigation models successfully handle path planning but fail at the final decision stage which indicates they are still missing the ``last centimeter'' capability required for real-world deployment.

\noindent \textbf{\textit{Findings4: Deficiencies in Dynamic Social Interaction.}} Performance on dynamic interaction tasks exposes significant limitations. \texttt{Human Follow Success} (\texttt{HFS}), the task completion rate, and \texttt{Human Follow Ratio} (\texttt{HFR}), the proportion of valid following, remain low across all configurations. The best \texttt{HFS} of 45.16\% achieved by Uni-NaVid on Carter degrades substantially on legged robots. This persistent violation of personal space constraints reveals a lack of social awareness and poses critical challenges for real-world deployment.



\begin{table*}[!t]
    \centering
    \caption{\textbf{Ablation study of language prompt strategies.} All metrics are reported in \%. Best results are bolded.}
    \label{tab:styletest}
    \footnotesize

    \renewcommand{\arraystretch}{0.92}
    \setlength{\tabcolsep}{4pt}

    \sisetup{
        detect-weight=true,
        detect-inline-weight=math,
        table-number-alignment=center
    }

    \begin{tabular}{ll *{10}{c}}  
    \toprule
    \textbf{Model} & \textbf{Variant} &
    {SR $\uparrow$} & {CSR $\uparrow$} & {SGC $\uparrow$} & {OSR $\uparrow$} &
    {SPL $\uparrow$} & {HFS $\uparrow$} & {HFR $\uparrow$} &
    {SII $\downarrow$} & {EQA-ACC $\uparrow$} \\
    \midrule
    
        \multirow{4}{*}{\textbf{OmniNav}}
        & Origin         & 8.74 & 1.94 & \textbf{35.30} & 15.53 & 7.01 & 14.29 & \textbf{17.49} & \textbf{3.40} & \multicolumn{1}{c}{--} \\
        & Concise        & 13.65 & 1.94 & 26.85 & \textbf{16.51} & 10.41 & 7.86 & 8.91 & 8.98 & \multicolumn{1}{c}{--} \\
        & First Person   & \textbf{14.47} & \textbf{3.88} & 25.02 & 14.32 & \textbf{11.88} & \textbf{25.72} & 13.06 & 13.18 & \multicolumn{1}{c}{--} \\
        & Verbose        & 8.65 & 2.94 & 25.09 & 9.47 & 6.69 & \textbf{25.72} & 11.88 & 7.88 & \multicolumn{1}{c}{--} \\
        \midrule
        \multirow{4}{*}{\textbf{Uni-NaVid}}
        & Origin         & 1.96 & 2.94 & \textbf{44.54} & 26.47 & 1.56 & 23.33 & \textbf{32.75} & 18.46 & 6.25 \\
        & Concise        & 4.55 & 1.82 & 38.74 & 25.45 & \textbf{3.72} & 24.24 & 27.85 & \textbf{14.46} & \textbf{10.53} \\
        & First Person   & \textbf{5.83} & 1.94 & 39.83 & \textbf{27.18} & 3.28 & \textbf{35.71} & 31.10 & 19.34 & 5.26 \\
        & Verbose        & 3.88 & \textbf{3.88} & 39.47 & 18.45 & 2.85 & 32.14 & 31.59 & 15.50 & 5.26 \\
        \bottomrule
    \end{tabular}
    \vspace{-1em}
\end{table*}



\noindent \textbf{\textit{Findings5: Sensitivity to Morphological Variations.}} Performance varies dramatically across robot morphologies even for identical algorithms. OmniNav achieves 8.74\% \texttt{SR} on the quadrupedal Aliengo but drops to 1.96\% on the wheeled Carter, while the humanoid platform consistently produces the lowest performance due to its complex dynamics. This sensitivity can be attributed to the inherent differences in embodiment complexity where distinct kinematic constraints influence trajectory feasibility and the increased control difficulty of bipedal locomotion introduces greater physical instability compared to wheeled or quadrupedal platforms.

\begin{figure}[htbp] 
    \centering
    \includegraphics[width=\linewidth]{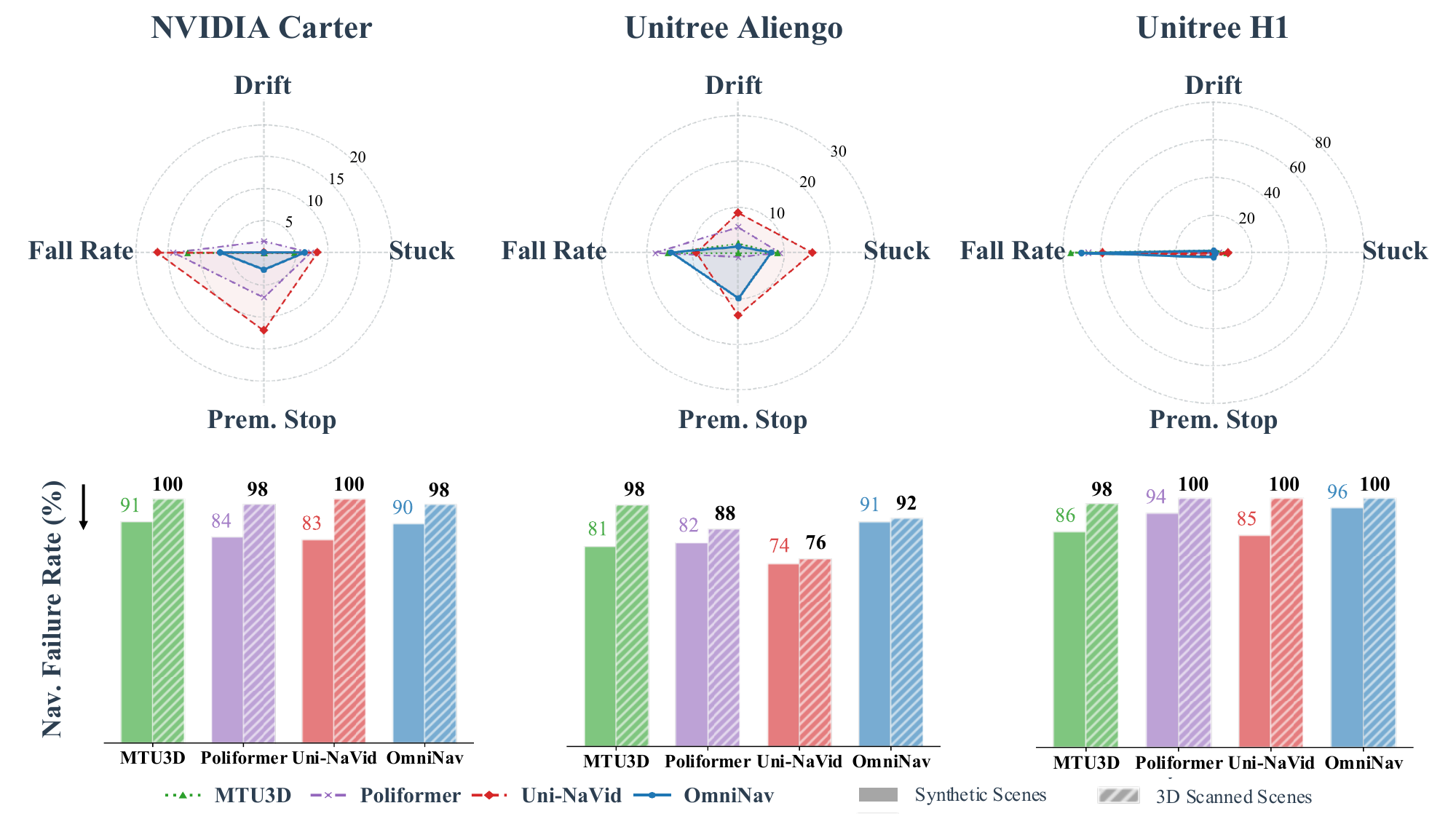} 
    \caption{\textbf{Failure Mode Analysis across Embodiments.} Radar charts comparing four failure types for four models. Bar chats comparing total failure on two scenes.}
    \label{fig:failure_analysis}
    \vspace{-2em}
\end{figure}



\subsection{Failure Mode Analysis}
To provide a more granular characterization of the performance bottlenecks, we categorize the observed failures into four primary modes (Fig. \ref{fig:failure_analysis} top): (i) \texttt{Goal-reaching Drift} (\texttt{Drift}), where the agent reaches the target but leaves again; (ii) \texttt{Termination Signal Failure} (\texttt{Stuck}), where the agent reaches the goal but fails to claim completion; (iii) \texttt{Premature Termination} (\texttt{Prem.~Stop}), involving episodes that stop before reaching the target vicinity; (iv) \texttt{Fall}, which accounts for failures resulting from falling. To further disentangle the failure patterns, we extend analysis to the interplay between scenario types and task performance, investigating how the structural complexity of different environments influences the distribution of failure modes.

\noindent \textbf{\textit{Findings6: Failure Modes Various across Embodiments.}} The distribution of failure modes differs substantially across robot morphologies (Fig. \ref{fig:failure_analysis} top). On the wheeled Carter, \texttt{Stuck} and \texttt{Prem.~Stop} constitute the dominant failure modes, while \texttt{Fall} remains minimal. Falls on the wheeled Carter are caused by collisions rather than balance loss. In contrast, the humanoid H1 exhibits an elevated rate of \texttt{Fall} (60--80\%), which suppresses the relative proportion of other failure types. The quadrupedal Aliengo presents a more balanced distribution. We attribute this divergence to the inherent physical properties and control stability margins of each embodiment. The wheeled platform possesses a statically stable base, meaning that policy uncertainty or sub-optimal actions typically manifest passively as hesitation or idling (\texttt{Stuck}, \texttt{Prem.~Stop}). Conversely, the bipedal H1 requires continuous active balancing and is highly sensitive to control noise. Consequently, similar policy imperfections propagate into dynamic instability, causing immediate falls rather than benign stops. Thus, the variation in failure modes reflects how distinct dynamic constraints amplify different consequences of the same policy limitations.

\noindent \textbf{\textit{Findings7: Failure Modes Between Synthetic and Real Scenes.}} A comparative analysis (Fig. \ref{fig:failure_analysis} bottom) between synthetic environments and 3D scanned scenes  reveals a consistent performance degradation within photorealistic reconstructions.  \texttt{Failure Rate} is defined as $1 - \texttt{SR}$. Across all model-embodiment configurations, Matterport3D scenes induce significantly higher failure rates than their synthetic counterparts. We attribute this decline to the inherent structural complexity of real-world scans, characterized by irregular geometries, dense clutter, and narrow passages, which impose stricter demands on both perceptual understanding and precise kinematic planning. Consequently, while synthetic scenes with predictable layouts tend to mask policy limitations, the unstructured nature of real-world environments exposes the models' fragility and highlights the critical gap between simplified training proxies and realistic deployment scenarios.

\subsection{Instruction Style Analysis}
\noindent \textbf{\textit{Findings8: Sensitivity to Instruction Variations.}} To investigate the impact of linguistic variability on model robustness, we conducted systematic tests on the OmniNav and Uni-NaVid using the Unitree Aliengo platform. The results in Tab.~\ref{tab:styletest} reveal that navigation performance is highly sensitive to instruction phrasing, with distinct styles eliciting varied responses across different metrics. Notably, the \textit{Concise} and \textit{First Person} variants consistently improve \texttt{Success Rate} (\texttt{SR}) compared to the original instructions for both models, whereas the \textit{Verbose} style yields comparable or degraded performance. Furthermore, while \texttt{SGC} and \texttt{HFR} consistently decrease across all stylized variants, \texttt{HFS} peaks specifically under the \textit{First Person} style. This observed sensitivity suggests that current models overfit to specific syntactic patterns and struggle to generalize across diverse linguistic formulations, highlighting a critical need for greater linguistic diversity during training to achieve robust natural language understanding.


\section{Conclusion} 
\label{sec:conclusion}

We present OmniNavBench, a comprehensive platform that bridges the divide between specialized navigation primitives and general-purpose embodiment. Through the integration of composite instructions and diverse robot morphologies, we provide a robust testbed for cross-skill and cross-embodiment generalization. Crucially, our analysis reveals a systemic failure in current unified models to handle sequential skill transitions. These findings underscore two critical directions: robust progress tracking for sequential coordination, and reliable termination decisions to close the gap between goal reaching and task completion. More broadly, our results highlight urgent challenges in dynamic interaction, morphological adaptability, and instruction robustness, underscoring the need for next-generation agents capable of long-horizon reasoning.

\textbf{Acknowledgement.} This work was supported by the Natural Science Foundation of China (Grant No. 62503323).




\bibliographystyle{plainnat}
\bibliography{references}

\clearpage
\appendix

\subsection{Instruction Generation Pipeline}
The generation of a single episode in OmniNavBench follows a structured pipeline to ensure the synchronization of environment states, task logic, and expert behavior:

\begin{enumerate}
\item \textbf{Environment and State Initialization}: A human annotator performs a semantic traversal of the selected scene to identify a valid initial pose $s_0$ for the robot. During this stage, the annotator also manually places dynamic obstacles and defines their motion constraints $\mathcal{C}$ within the Isaac Sim environment.

\item \textbf{Task Orchestration}: A composite mission is constructed by sequencing at least two heterogeneous sub-tasks into a task chain $\mathcal{T} = \{(l_{1},\tau_{1},g_{1}), \dots, (l_{n},\tau_{n},g_{n})\}$. This includes defining the semantic goals $g_i$ and the corresponding natural language instruction segments $l_i$.

\item \textbf{Expert Demonstration}: A human operator teleoperates the selected robot morphology to execute the task chain $\mathcal{T}$ within the initialized simulation state. The system synchronously records egocentric RGB-D streams, robot poses, and control commands at each timestep.

\item \textbf{Verification and Post-processing}: The collected trajectory is audited via a verification loop where experts review the egocentric video to confirm that all sub-goals $g_i$ were successfully reached and that social constraints were maintained. Validated episodes are then aggregated into the final experience dataset $\mathcal{D}$.

\end{enumerate}

\subsection{Instruction Stylization Pipeline}

We developed a instruction stylization pipeline. Building upon the 1,779 core expert-annotated instructions, we utilized the Qwen3-Max~\cite{qwen3} large language model to generate semantically consistent paraphrases across three distinct linguistic dimensions: \textit{Concise}, \textit{Verbose}, and \textit{First Person}. This process yielded a dataset of 7,116 stylized instructions.

\begin{figure}[t]
  \centering
  \includegraphics[width=\linewidth]{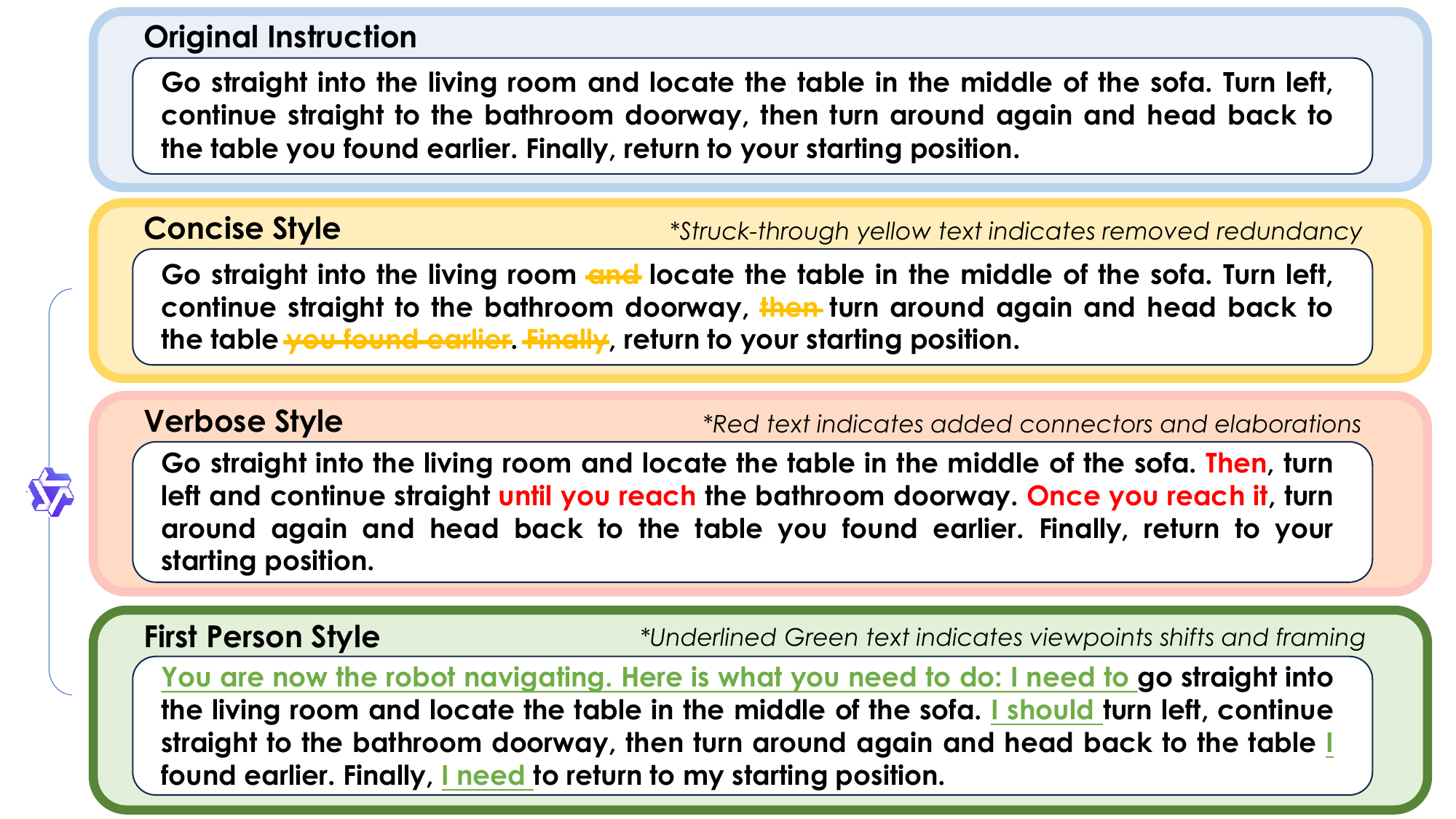} 
  
 \caption{\textbf{Semantically Consistent Instruction Style Generation Pipeline.} The figure illustrates the process of transforming a human-annotated original instruction into three distinct variants (\textit{Concise}, \textit{Verbose}, and \textit{First Person}) using Qwen3-Max. Visual annotations highlight specific linguistic modifications: removed redundancy (Concise), added elaborations (Verbose), and viewpoint shifts (First Person).}
  \label{fig:style_pipeline}
\end{figure}

The three variants were specifically designed to evaluate the robustness of embodied agents against natural language ambiguity, example shown in Fig. \ref{fig:style_pipeline}. All three variants are accompanied by manual verification to preserve the original navigation targets and action sequences, differing only in linguistic style:

\begin{itemize}
    \item \textbf{Concise}: Retains only core navigation intent by removing redundant modifiers and descriptions.
    \item \textbf{Verbose}: Adds environmental details and descriptive adjectives.
    \item \textbf{First Person}: Rewrites instructions from the agent's perspective.
\end{itemize}

\section{Experimental Settings}
\label{sec:exp_settings}

\subsubsection{Robot Embodiment Specifications}
The specific parameters for the different robot embodiments used in our ``Robot Pool'' are detailed in Table~\ref{tab:robots}. \texttt{Height} refers to the overall robot height. \texttt{Cam (Stand)} denotes the camera height in the default standing pose, while \texttt{Cam (Obs.)} indicates the camera height during active navigation, which may differ due to postural changes.

\begin{table}[h]
    \centering
    \caption{Specifications of Robot Embodiments}
    \small
    \label{tab:robots}
    \begin{tabular}{lccc}
        \toprule
        \textbf{Robot Type} & \textbf{Height} & \textbf{Cam (Stand)} & \textbf{Cam (Obs.)} \\
        & (m) & (m) & (m) \\
        \midrule
        Nvidia Carter V1  & 0.66 & 0.56 & 0.56 \\ 
        Unitree Aliengo   & 0.60 & 0.52 & 0.22 \\ 
        Unitree H1        & 1.80 & 1.47 & 1.36 \\ 
        \bottomrule
    \end{tabular}
\end{table}

\subsubsection{Simulation of Dynamic Humans}
\begin{figure}[htbp] 
    \centering
    \includegraphics[width=\linewidth]{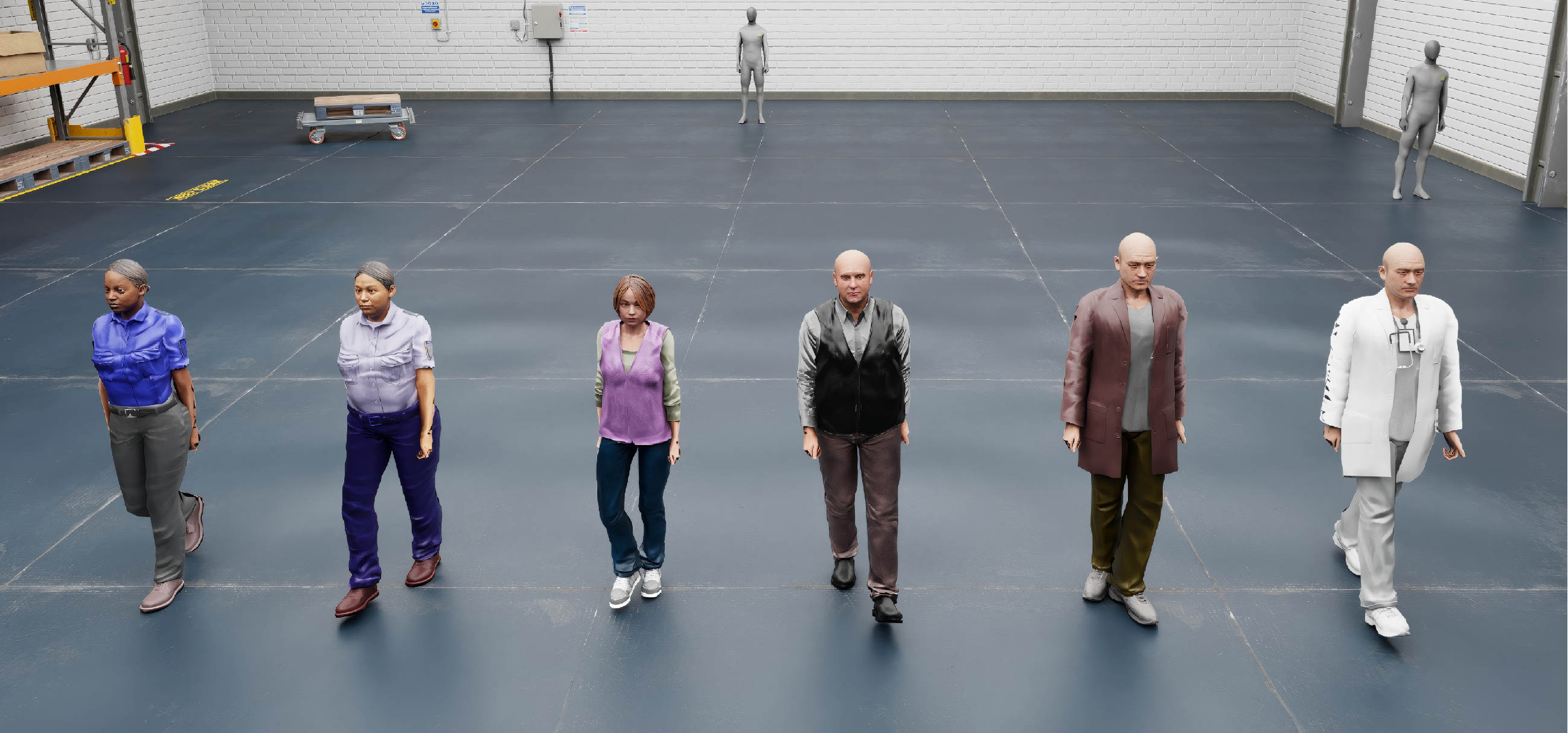} 
    \caption{\textbf{Dynamic human character appearances.} The six characters span diverse genders, ethnicities, and attire to improve visual diversity in Social Navigation and Human Following tasks.}
    \label{fig:characters}
\end{figure}

For Social Navigation and Human Following tasks, dynamic humans are simulated using the \texttt{omni.anim.people} extension in Isaac Sim. Human characters follow predefined waypoint sequences specified, which can be chained to form longer trajectories. All characters walk at a fixed speed of about 1.0\,m/s. Given the complexity of indoor spaces, each episode contains 1--2 dynamic humans, sampled from a pool of 6 distinct character appearances shown in Fig. \ref{fig:characters}. Trajectories are fully deterministic: replaying the same episode produces identical human motion, ensuring fair and reproducible evaluation.

\subsubsection{Metrics Calculation}

\textbf{Distance Computation}

\paragraph{Geodesic Distance via NavMesh.}
For VLN and ObjectNav tasks, we compute geodesic distance $d_{geo}(\cdot,\cdot)$ via Navigation Mesh (NavMesh). Existing ObjectNav evaluations typically measure the Euclidean distance between the robot and the object's geometric center. However, for large objects (\textit{e.g.}, beds, sofas), the geometric center often lies in unreachable areas, leading to false negatives. 

To address this, we leverage NavMesh, a polygonal representation of traversable areas in Isaac Sim. Starting from the target position, the system performs a fan-shaped scan toward the robot to identify the nearest reachable point on the NavMesh boundary, then computes the geodesic path distance.

\paragraph{Euclidean Distance.}
For PointNav, HumanFollow, and SocialNav, we directly compute the 2D Euclidean distance $d(\cdot,\cdot)$ between the robot's center $x_t$ and the target position.

\textbf{Metric Definitions}

\paragraph{Distance Function Assignment.}
Throughout the following metric definitions, the distance function $d(\cdot, \cdot)$ is instantiated according to task type:
\begin{itemize}
    \item For \textbf{VLN} and \textbf{ObjectNav} sub-tasks, $d(\cdot, \cdot) := d_{geo}(\cdot, \cdot)$ denotes the geodesic distance computed via NavMesh.
    \item For \textbf{PointNav}, \textbf{HumanFollow}, and \textbf{SocialNav} sub-tasks, $d(\cdot, \cdot)$ denotes the 2D Euclidean distance.
\end{itemize}
Let $\delta_g = 2.0\,\text{m}$ denote the default goal distance threshold for episode-level metrics.

\paragraph{Completion Metrics.}

\begin{itemize}
    \item \textbf{SR (Success Rate):} The fraction of episodes where the agent issues \texttt{Stop} within $\delta_g$ of the final goal.
    \[ \text{SR} = \frac{1}{N} \sum_{i=1}^{N} S_i(\delta_g) \]
    where $N$ is the total number of test episodes, and $S_i(\delta_g) = 1$ if the agent stops within distance $\delta_g$ of the final goal in episode $i$, otherwise $0$.

    \item \textbf{CSR (Completion Success Rate):} The average ratio of strictly completed sub-instructions per episode.
    \[ \text{CSR} = \frac{1}{N} \sum_{i=1}^{N} \frac{K_{\text{completed},i}}{K_{\text{total},i}} \]
    where $K_{\text{total},i}$ is the total number of sub-instructions in episode $i$, and $K_{\text{completed},i}$ is the number of sub-instructions completed according to task-specific rules (see Tab.~\ref{tab:task_definitions}).

\item \textbf{SGC (Sub-Goal Completion):} The average continuous progress across all sub-tasks.
    \[ \text{SGC} = \frac{1}{N} \sum_{i=1}^{N} \frac{1}{K_i} \sum_{j=1}^{K_i} \mathcal{P}(\tau_{i,j}) \]
    where $K_i$ is the number of sub-tasks in episode $i$, and $\tau_{i,j}$ denotes the $j$-th sub-task. The progress function $\mathcal{P}(\cdot)$ is defined according to sub-task type:
    
    \begin{itemize}
        \item \textit{State-Goal Navigation} (\texttt{GOTO\_OBJECT} for ObjectNav, \texttt{GOTO\_POINT} for PointNav, \texttt{GOTO\_LANDMARK} and \texttt{GOTO\_ROOM} as markers for VLN, \texttt{RETURN\_TO} for backtracking):
        \[
        \mathcal{P} = \min\left(1,\; \frac{r}{d_{\min}}\right)
        \]
        where $d_{\min}$ is the minimum distance to the sub-goal during the episode, and $r$ is the task-specific success threshold.
        
        \item \textit{Room Entry} (\texttt{GOTO\_ROOM}):
        \[
        \mathcal{P} = \mathbb{I}[\text{entered}]
        \]
        Binary indicator: 1 if the agent entered the target room, 0 otherwise.
        
        \item \textit{Human Following} (\texttt{FOLLOW\_HUMAN}):
        \[
        \mathcal{P} = \text{HFR}
        \]
        Directly uses the Human Follow Ratio as the progress measure.
    \end{itemize}

    \item \textbf{OSR (Oracle Success Rate):} Success if the agent was \textit{ever} within $\delta_g$ of the final goal during the episode.
    \[ \text{OSR} = \frac{1}{N} \sum_{i=1}^{N} OS_i(\delta_g) \]
    where $OS_i(\delta_g) = 1$ if the agent reaches within $\delta_g$ of the final goal at any timestep, regardless of whether it issued \texttt{Stop}.
\end{itemize}

\paragraph{Efficiency Metrics.}

\begin{itemize}
    \item \textbf{SPL (Success weighted by Path Length)~\cite{anderson2018evaluation}:} Measures navigation efficiency, weighted by success.
    \[ \text{SPL} = \frac{1}{N} \sum_{i=1}^{N} S_i \frac{\ell_{\text{shortest},i}}{\max(\ell_{\text{actual},i}, \ell_{\text{shortest},i})} \]
    where $\ell_{\text{shortest},i}$ is the shortest path length from start to goal, and $\ell_{\text{actual},i}$ is the actual path length traversed by the agent.
\end{itemize}

\paragraph{Interaction Metrics.}

\begin{itemize}
    \item \textbf{HFS (Human Follow Success):} Adopted form \cite{wang2025trackvla}, we defined binary success at the final timestep $t_e$ when the target human stops.
    \[ \text{HFS} = \frac{1}{N_{\text{follow}}} \sum_{i=1}^{N_{\text{follow}}} \mathbb{I}(\text{Valid}_{i,t_e}) \]
    where $N_{\text{follow}}$ is the number of HumanFollow episodes. $d_t = d(x_t, h_t)$ denote the Euclidean distance between the robot's  center $x_t$ and the target human position $h_t$ at timestep $t$, and $\angle_{err,t}$ denote the angular difference between the robot's heading and the direction toward the target human. The valid following condition is defined as:
    \[ \text{Valid}_{i,t} := (d_t \in (1.0, 3.0]\,\text{m}) \land (\angle_{err,t} \le 60^\circ) \]

    \item \textbf{HFR (Human Follow Ratio):} Adopted form \cite{wang2025trackvla}, we defined the proportion of timesteps satisfying the valid following condition.
    \[ \text{HFR} = \frac{1}{N_{\text{follow}}} \sum_{i=1}^{N_{\text{follow}}} \frac{T_{\text{valid},i}}{T_{\text{follow},i}}, \quad T_{\text{valid},i} = \sum_{t=1}^{T_{\text{follow},i}} \mathbb{I}(\text{Valid}_{i,t}) \]
    where $T_{\text{follow},i}$ is the total duration of the following phase in episode $i$.

   \item \textbf{EQA-ACC (EQA Accuracy):} Adopted form \cite{zhang2024uni}, we defined accuracy of embodied question answering.
    \[ \text{EQA-ACC} = \frac{1}{N_{\text{eqa}}} \sum_{i=1}^{N_{\text{eqa}}} \mathbb{I}(a_{\text{pred},i} \cong a_{\text{gt},i}) \]
    where $N_{\text{eqa}}$ is the number of EQA episodes, $a_{\text{pred},i}$ and $a_{\text{gt},i}$ are the predicted and ground-truth answers. The matching operator $\cong$ performs case-insensitive substring matching, with numeric normalization that unifies written numerals (\textit{e.g.}, ``one'', ``two'') and digits (\textit{e.g.}, ``1'', ``2'') before comparison.
\end{itemize}

\paragraph{Safety Metrics.}

\begin{itemize}
    \item \textbf{SII (Social Intrusion Index):} The ratio of timesteps where the robot violates personal space.
    \[ \text{SII} = \frac{1}{N_{\text{social}}} \sum_{i=1}^{N_{\text{social}}} \left( \frac{1}{T_i} \sum_{t=1}^{T_i} \mathbb{I}(d(x_t, \mathcal{H}) < 1.2\,\text{m}) \right) \]
    where $N_{\text{social}}$ is the number of episodes involving dynamic humans, $T_i$ is the episode length, $x_t$ is the robot's center position, $\mathcal{H}$ is the set of humans, and $d(\cdot,\cdot)$ is the Euclidean distance.
\end{itemize}

\subsection{Sim-to-Real Validation}
\label{sec:sim2real}

\begin{figure}[t]
    \centering
    \includegraphics[width=\linewidth]{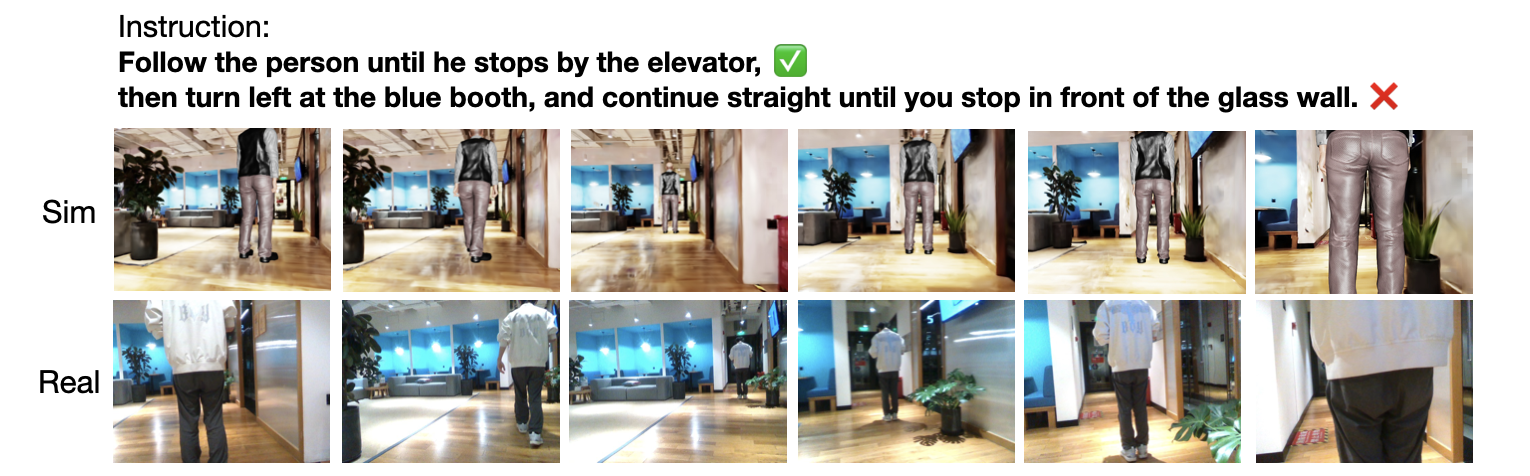}
    \caption{Sim vs.\ Real validation. Same instructions, approximately consistent completion in each setting.}
    \label{fig:simReal}
\end{figure}

To validate sim-to-real consistency, we deployed Uni-NaVid on a wheeled robot in a real-world scene and its 3DGS-reconstructed simulated counterpart. As shown in Fig.~\ref{fig:simReal}, the agent exhibits qualitatively consistent behaviors across both settings. Moreover, OmniNavBench enables comprehensive evaluation by encompassing both purely synthetic environments and imported real-world scans like Matterport3D, providing fair, reproducible, and safe evaluations across tasks and embodiments.

\subsection{Behavioral Analysis During Pedestrian Pauses}
\label{sec:hesitation}

Unlike human operators, who naturally pause to assess unpredictable situations (\textit{e.g.}, verifying if a pedestrian has stopped), current models lack this cognitive ``hesitation''. To quantify this gap, we analyze agent behaviors during pedestrian pause windows (Tab.~\ref{tab:hesitation}) and classify them into four states: \emph{Correct Wait}, \emph{Abandon}, \emph{Social Intrusion}, and \emph{Overshoot}. Our findings reveal that existing algorithms struggle significantly to interpret unpredictable dynamic behaviors. Rather than exhibiting the \emph{Correct Wait} behavior seen in ground-truth (GT) demonstrations, models typically either prematurely \emph{Abandon} the target (degrading HFR) or blindly move forward, resulting in \emph{Social Intrusion} (worsening SII).

\begin{table}[h]
\centering
\caption{Behavioral classification during pedestrian pause windows.}
\label{tab:hesitation}
\scriptsize
\renewcommand{\arraystretch}{0.8}
\resizebox{0.85\linewidth}{!}{%
\begin{tabular}{lcccccc}
\toprule
\textbf{Agent} & \textbf{Correct Wait}\,$\uparrow$ & \textbf{Abandon}\,$\downarrow$ & \textbf{Intrusion}\,$\downarrow$ & \textbf{Overshoot}\,$\downarrow$ & \textbf{HFR}\,$\uparrow$ & \textbf{SII}\,$\downarrow$ \\
\midrule
GT Operator & \textbf{81.8} & \textbf{2.3} & 15.9 & \textbf{0.0} & -- & -- \\
Uni-NaVid & 29.4 & 29.4 & 35.3 & 5.9 & \textbf{32.75} & 34.28 \\
OmniNav & 37.5 & 62.5 & 0.0 & 0.0 & 17.49 & 23.49 \\
MTU3D & 25.0 & 62.5 & \textbf{12.5} & 0.0 & 16.91 & 19.48 \\
Poliformer & 13.3 & 86.7 & 0.0 & 0.0 & 9.68 & \textbf{14.81} \\
\bottomrule
\end{tabular}}
\end{table}

\subsection{Instruction Statistics}
\begin{figure}[t] 
    \centering
    \includegraphics[width=\linewidth]{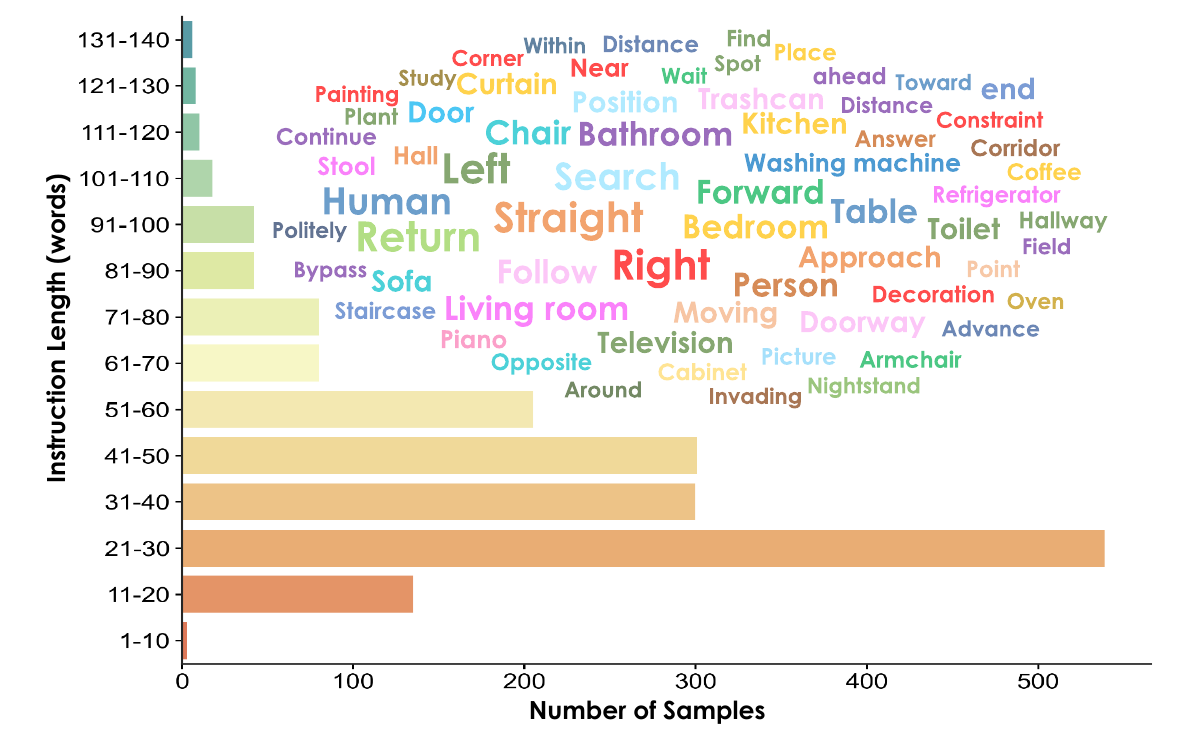} 
    \caption{\textbf{Statistics of instructions.} 
    Distribution of instruction lengths and a word cloud of navigation targets.}
    \label{fig:length_appendix}
\end{figure}

\subsection{Assets}
\label{sec:assets}

OmniNavBench is constructed using 170 3D scenes from two primary sources: Matterport3D (MP3D)~\cite{chang2017matterport3d} and GRScenes~\cite{wang2024grutopia}. Tab.~\ref{tab:scene_stats} summarizes the scene distribution.

\begin{table}[h]
\centering
\caption{Scene statistics used in OmniNavBench.}
\label{tab:scene_stats}
\begin{tabular}{lc}
\toprule
\textbf{Dataset} & \textbf{Number of Scenes} \\
\midrule
Matterport3D (MP3D) & 85 \\
GRScenes-Home & 61 \\
GRScenes-Commercial & 24 \\
\midrule
\textbf{Total} & \textbf{170} \\
\bottomrule
\end{tabular}
\end{table}

\subsubsection{Matterport3D Scenes}
We utilize 85 scenes from the Matterport3D dataset. The complete list of scene identifiers is provided below:

\par\noindent\footnotesize\ttfamily
17DRP5sb8fy, 1LXtFkjw3qL, 1pXnuDYAj8r, 29hnd4uzFmX, 2azQ1b91cZZ, 2n8kARJN3HM, 2t7WUuJeko7, 5LpN3gDmAk7, 5q7pvUzZiYa, 759xd9YjKW5, 7y3sRwLe3Va, 8194nk5LbLH, 82sE5b5pLXE, 8WUmhLawc2A, ARNzJeq3xxb, B6ByNegPMKs, D7N2EKCX4Sj, E9uDoFAP3SH, EDJbREhghzL, EU6Fwq7SyZv, GdvgFV5R1Z5, HxpKQynjfin, JF19kD82Mey, JeFG25nYj2p, JmbYfDe2QKZ, PX4nDJXEHrG, Pm6F8kyY3z2, PuKPg4mmafe, QUCTc6BB5sX, RPmz2sHmrrY, S9hNv5qa7GM, SN83YJsR3w2, TbHJrupSAjP, ULsKaCPVFJR, UwV83HsGsw3, Uxmj2M2itWa, V2XKFyX4ASd, VFuaQ6m2Qom, VLzqgDo317F, VVfe2KiqLaN, Vt2qJdWjCF2, Vvot9Ly1tCj, VzqfbhrpDEA, WYY7iVyf5p8, X7HyMhZNoso, XcA2TqTSSAj, YFuZgdQ5vWj, YVUC4YcDtcY, YmJkqBEsHnH, Z6MFQCViBuw, ZMojNkEp431, aayBHfsNo7d, ac26ZMwG7aT, b8cTxDM8gDG, cV4RVeZvu5T, e9zR4mvMWw7, fzynW3qQPVF, gTV8FGcVJC9, gYvKGZ5eRqb, gxdoqLR6rwA, i5noydFURQK, jh4fc5c5qoQ, jtcxE69GiFV, kEZ7cmS4wCh, mJXqzFtmKg4, oLBMNvg9in8, p5wJjkQkbXX, pRbA3pwrgk9, pa4otMbVnkk, q9vSo1VnCiC, qoiz87JEwZ2, r1Q1Z4BcV1o, r47D5H71a5s, rPc6DW4iMge, rqfALeAoiTq, s8pcmisQ38h, sKLMLpTHeUy, sT4fr6TAbpF, uNb9QFRL6hY, ur6pFq6Qu1A, vyrNrziPKCB, wc2JMjhGNzB, x8F5xyUWy9e, yqstnuAEVhm, zsNo4HB9uLZ
\normalfont\normalsize\par

\subsubsection{GRScenes-Home Scenes}
We utilize 61 residential scenes from GRScenes. The scene identifiers are:

\par\noindent\footnotesize\ttfamily
MV7J6NIKTKJZ2AABAAAAADI8, MV7J6NIKTKJZ2AABAAAAADQ8, MV7J6NIKTKJZ2AABAAAAADY8, MV7J6NIKTKJZ2AABAAAAAEA8, MV7J6NIKTKJZ2AABAAAAAEI8, MVUCSQAKTKJ5EAABAAAAAAA8, MVUCSQAKTKJ5EAABAAAAAAI8, MVUCSQAKTKJ5EAABAAAAAAQ8, MVUCSQAKTKJ5EAABAAAAAAY8, MVUCSQAKTKJ5EAABAAAAABA8, MVUCSQAKTKJ5EAABAAAAABI8, MVUCSQAKTKJ5EAABAAAAABQ8, MVUCSQAKTKJ5EAABAAAAABY8, MVUCSQAKTKJ5EAABAAAAACA8, MVUCSQAKTKJ5EAABAAAAACI8, MVUCSQAKTKJ5EAABAAAAACQ8, MVUCSQAKTKJ5EAABAAAAADI8, MVUCSQAKTKJ5EAABAAAAADQ8, MVUCSQAKTKJ5EAABAAAAADY8, MVUCSQAKTKJ5EAABAAAAAEI8, MVUHL5YKTKJ5EAABAAAAAAI8, MVUHL5YKTKJ5EAABAAAAAAQ8, MVUHLWYKTKJ5EAABAAAAAAA8, MVUHLWYKTKJ5EAABAAAAAAI8, MVUHLWYKTKJ5EAABAAAAAAQ8, MVUHLWYKTKJ5EAABAAAAAAY8, MVUHLWYKTKJ5EAABAAAAABA8, MVUHLWYKTKJ5EAABAAAAABI8, MVUHLWYKTKJ5EAABAAAAABQ8, MVUHLWYKTKJ5EAABAAAAABY8, MVUHLWYKTKJ5EAABAAAAACA8, MVUHLWYKTKJ5EAABAAAAACQ8, MVUHLWYKTKJ5EAABAAAAACY8, MVUHLWYKTKJ5EAABAAAAADA8, MVUHLWYKTKJ5EAABAAAAADI8, MVUHLWYKTKJ5EAABAAAAADQ8, MVUHLWYKTKJ5EAABAAAAAEA8, MVUHLWYKTKJ5EAABAAAAAEI8, MWAX5JYKTKJZ2AABAAAAAAA8, MWAX5JYKTKJZ2AABAAAAAAI8, MWAX5JYKTKJZ2AABAAAAAAY8, MWAX5JYKTKJZ2AABAAAAABA8, MWAX5JYKTKJZ2AABAAAAABI8, MWAX5JYKTKJZ2AABAAAAABQ8, MWAX5JYKTKJZ2AABAAAAABY8, MWAX5JYKTKJZ2AABAAAAACA8, MWAX5JYKTKJZ2AABAAAAADA8, MWAX5JYKTKJZ2AABAAAAADI8, MWAX5JYKTKJZ2AABAAAAADQ8, MWAX5JYKTKJZ2AABAAAAADY8, MWAX5JYKTKJZ2AABAAAAAEA8, MWAX5JYKTKJZ2AABAAAAAEI8, MWBGLKQKTKJZ2AABAAAAAAA8, MWBGLKQKTKJZ2AABAAAAAAI8, MWBGLKQKTKJZ2AABAAAAAAQ8, MWBGLKQKTKJZ2AABAAAAAAY8, MWBGLKQKTKJZ2AABAAAAABA8, MWBGLKQKTKJZ2AABAAAAABI8, MWBGLKQKTKJZ2AABAAAAABQ8, MWBGLKQKTKJZ2AABAAAAABY8, MWBGLKQKTKJZ2AABAAAAACA8
\normalfont\normalsize\par

\subsubsection{GRScenes-Commercial Scenes}
We utilize 24 commercial scenes from GRScenes. The scene identifiers are:

\par\noindent\footnotesize\ttfamily
MV4AFHQKTKJZ2AABAAAAAEA8, MV4AFHQKTKJZ2AABAAAAAEI8, MV5M25QKTKJZ2AABAAAAAAA8, MV5M25QKTKJZ2AABAAAAAAI8, MV5M25QKTKJZ2AABAAAAAAQ8, MV5M25QKTKJZ2AABAAAAAAY8, MV5M25QKTKJZ2AABAAAAAEI8, MV7J6NIKTKJZ2AABAAAAAAA8, MV7J6NIKTKJZ2AABAAAAAAI8, MVJWVGYKTLDAYAABAAAAAAQ8, MVSGSAIKTKJ66AABAAAAAEA8, MVSYCXYKTKJ66AABAAAAAAA8, MVSYCXYKTKJ66AABAAAAACY8, MVSYCXYKTKJ66AABAAAAADA8, MVSYCXYKTKJ66AABAAAAADI8, MWF4WLIKTIFZIAABAAAAABY8, MWF4WLIKTIFZIAABAAAAACA8, MWF4WLIKTIFZIAABAAAAACI8, MWF4WLIKTIFZIAABAAAAACQ8, MWF4WLIKTIFZIAABAAAAADA8, MWF4WLIKTIFZIAABAAAAADI8, MWF4WLIKTIFZIAABAAAAAEA8, MWF4WLIKTIFZIAABAAAAAEI8, MWHLEPQKTIFZIAABAAAAAAA8
\normalfont\normalsize\par

\end{document}